\documentclass[journal]{IEEEtran}

\ifCLASSINFOpdf
\usepackage[pdftex]{graphicx}
\graphicspath{{./}}

\DeclareGraphicsExtensions{.png,.pdf,.jpg,.jpeg}
\else
   \usepackage[dvips]{graphicx}
   \graphicspath{{Figures/}}
     \DeclareGraphicsExtensions{.png,.pdf,.jpg,.jpeg}
\fi

\usepackage{arydshln} 

\usepackage{pifont}
\usepackage{amsmath,amssymb,amsfonts}
\usepackage{bm}
\usepackage{graphicx}
\usepackage{mathtools}
\usepackage{array,ragged2e}
\usepackage{booktabs}
\usepackage{makecell}
\usepackage{threeparttable}
\usepackage{color}
\usepackage[table,xcdraw,dvipsnames]{xcolor}
\usepackage{diagbox}
\definecolor{myPink}{rgb}{0.9294, 0.0078, 0.5490}
\definecolor{Gray}{gray}{0.9}
\definecolor{mycolor}{HTML}{FBEAFF}  
\definecolor{xx}{HTML}{00C9A7}
\usepackage[
  colorlinks=true,
  linkcolor=red,  
  citecolor=green,  
  urlcolor=myPink,   
  linkbordercolor={1 1 1} 
]{hyperref}
\usepackage{threeparttable}
\usepackage{diagbox}
\usepackage{orcidlink}

\makeatletter
\def\url@leostyle{%
  \@ifundefined{selectfont}{\def\UrlFont{\sf}}{\def\UrlFont{\small\ttfamily}}}
\makeatother

\usepackage{fancyhdr}

\usepackage{amsmath,amsfonts}
\usepackage{algorithmic}
\usepackage{algorithm}
\usepackage{array}
\usepackage[caption=false,font=normalsize,labelfont=sf,textfont=sf]{subfig}
\usepackage{textcomp}
\usepackage{stfloats}
\usepackage{url}
\usepackage{verbatim}
\usepackage{graphicx}
\usepackage{cite}

\usepackage{multirow}

\makeatletter
\let\NAT@parse\undefined
\makeatother
\usepackage{hyperref}

\begin{document}

\title{Joint multi-dimensional dynamic attention and transformer for general image restoration}


\author{Huan~Zhang,~\IEEEmembership{Member,~IEEE,}
        Xu~Zhang*,
	    Nian~Cai*,~\IEEEmembership{Member,~IEEE,}   
	    Jianglei~Di,~\IEEEmembership{Member,~IEEE,} 
     and Yun~Zhang,~\IEEEmembership{Senior Member,~IEEE}
	    

\thanks{This work was in part supported by xxx. \emph{(Corresponding author: Xu Zhang; Nian Cai.)}}%

\thanks{Huan Zhang, Xu Zhang, and Nian Cai are with the School of Information Engineering, Guangdong University of Technology, Guangzhou 510006, China (e-mail: huanzhang2021@gdut.edu.cn; 2122103221@mail2.gdut.edu.cn; cainian@gdut.edu.cn).}%

\thanks{Jianglei Di is with Key Laboratory of Photonic Technology for Integrated Sensing and Communication, Ministry of Education, and Guangdong Provincial Key Laboratory of Information Photonics Technology, Institute of Advanced Photonics Technology, School of Information Engineering, Guangdong University of Technology, Guangzhou 510006, China. (e-mail: jiangleidi@gdut.edu.cn).}

\thanks{Yun Zhang is with the School of Electronics and Communication Engineering, Sun Yat-Sen University, Shenzhen 518107, China (e-mail: zhangyun2@mail.sysu.edu.cn).}

    }

\markboth{Journal of \LaTeX\ Class Files,~Vol.~14, No.~8, August~2021}%
{Shell \MakeLowercase{\textit{et al.}}: A Sample Article Using IEEEtran.cls for IEEE Journals}


\maketitle

\begin{abstract}

Outdoor images often suffer from severe degradation due to rain, haze, and noise, impairing image quality and challenging high-level tasks. Current image restoration methods struggle to handle complex degradation while maintaining efficiency. This paper introduces a novel image restoration architecture that combines multi-dimensional dynamic attention and self-attention within a U-Net framework. To leverage the global modeling capabilities of transformers and the local modeling capabilities of convolutions, we integrate sole CNNs in the encoder-decoder and sole transformers in the latent layer. Additionally, we design convolutional kernels with selected multi-dimensional dynamic attention to capture diverse degraded inputs efficiently. A transformer block with transposed self-attention further enhances global feature extraction while maintaining efficiency. Extensive experiments demonstrate that our method achieves a better balance between performance and computational complexity across five image restoration tasks: deraining, deblurring, denoising, dehazing, and enhancement, as well as superior performance for high-level vision tasks. The source code will be available at \url{https://github.com/House-yuyu/MDDA-former}.
\end{abstract}

\IEEEpeerreviewmaketitle

\section{Introduction}

\IEEEPARstart{I}{n} recent years,
the development of artificial intelligence and deep learning has led to significant advances in image restoration. However, captured images often suffer from various degradations due to changing weather conditions and sensor noise, such as haze, rain, sensor noise, blur, or low light. These degradations can affect the precise perception of the surroundings and impact subsequent applications like object detection \cite{GAUTAM2023103648} and tracking. To prompt more precise image processing tasks for safety, image restoration is highly desired. 


	
During the past decades, for physical causes of various degradations are different, specific and strong image priors have been tailored for respective image restoration. With the development of deep learning, which could directly learn complex mapping rules between degraded and clean images from large-scale data, the performance is significantly improved compared with traditional restoration approaches. Recently, convolutional neural networks (CNN) has been successfully employed for handling different degradations
due to its ability of learning generalizability, such as residual learning \cite{DnCNN,RDN,DuRN}, attention mechanisms \cite{BANet,FFA-Net}. 
{However, intricate degradation patterns of rain, haze, etc. require for large capacity of CNN, which leads to the increasing complexity that grows with the size of the network. Recently, dynamic convolution \cite{CondConv,DyConv} has been proposed to mitigate this problem by learning kernel-wise attention for convolutional filters. In addition, CNN-based methods have inherent flaws in capturing long-range dependencies in images.}

\begin{figure}[tp]
	\centerline{\includegraphics[page=1,trim = 0mm 0mm 0mm 0mm, clip, width=1\linewidth]{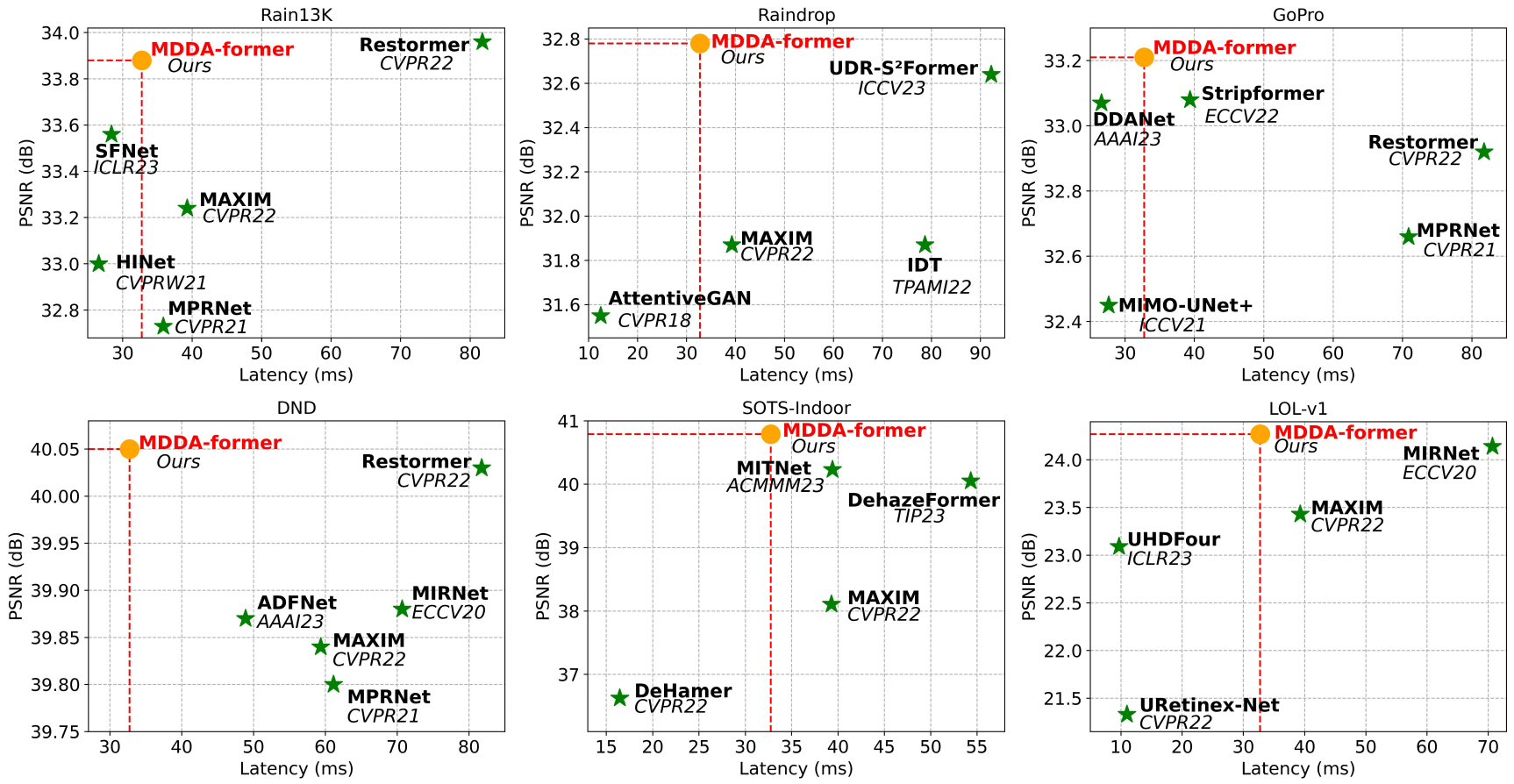}}
	\vspace{-1em}
	\caption{The performance/latency comparisons among our method and the representative state-of-the-art methods on six datasets across five image restoration tasks. } 
	\label{fig:performence}
	\vspace{-1.8em}
\end{figure}

Since high-level vision tasks have seen great progress propelled by the self-attention (SA) \cite{SA} which could capture long-range dependency within images, Transformer-based methods have shown great potential in the image restoration (IR) field. Despite of transformer’s good performance in IR, its computational complexity increases quadratically with image resolution. To alleviate this problem, Restormer \cite{Restormer} introduces a novel channel-wise self-attention and gated-Dconv feed-forward network, achieving approximately linear computational complexity. Uformer decreases the computational complexity within the window by leveraging a new locally-enhanced window attention mechanism \cite{Uformer}. 
Although the above Transformer-based methods utilize different schemes to decrease computational complexities, they still cost a large number of FLOPs and has slow inference speed due to the transformer architecture, such as 140.99G-81.76ms for Restormer and 89.46G-49.08ms for Uformer with regards to FLOPs and latency tested on images with 256$ \times $256 resolution. Compared with the CNN architecture, it is of lower efficiency to extract local information via the transformer architecture, which is performed at shallower layers \cite{Like}.

Recently, Transformer blocks combined with CNNs have been designed in the U-Net \cite{UNet} architecture to seek a good balance between performance and computational complexity in various tasks. Zhang \emph{et al.} \cite{SCUNet} introduced hybrid swin-conv blocks and embedded them into all the layers of U-shaped architecture for image denoising. Zhang \emph{et al.} \cite{AFD-former} proposed AFD-former by plugging asymmetric flow division unit into CNN-Transformer block for synthesized view quality enhancement. These works utilize the hybrid CNN-Transformer block in all the layers of U-shaped architectures to exploit CNN's local modeling ability and transformer's global modeling ability while achieving relatively low complexity. However, the effective and efficient way of where and how CNNs and transformers integrate into the U-shaped architecture needs further exploration.


In the U-shaped architecture, the latent layer (the deepest layer) has the most compact representation which {is beneficial} for SA to extract global features while the other layers present multi-scale early feature representation suitable for CNN to extract local information. {In view of this, there have been some recent works attempting to use CNNs for the encoder-decoder and transformers for the latent layer within UNet structure in other vision tasks, such as in image generation \cite{rombach2022high} and remote sensing image detection \cite{li2022transunetcd}}. Inspired by these, we propose a general image restoration architecture termed MDDA-former by equipping transformer blocks in the latent layer and CNN-based modules in the encoder-decoder explicitly, which can fully exploit multi-scale structural differences of the U-shaped architecture. {Furthermore, to improve the capacity of CNN blocks without sacrificing much efficiency,} a CNN-based Multi-Dimensional Dynamic Attention Block (MDAB) is designed in the encoder-decoder to learn dynamic complementary attentions of convolutional kernels and to perform local modeling effectively, which integrates spatial-wise, channel-wise, and filter-wise attentions. The Effective Transformer Block (ETB) in the latent layer is designed to capture global contextual information, which cascades a transposed attention \cite{Restormer} with a linear complexity and depth-wise convolution to improve the feature representation ability.

The contributions of this work can be summarized as follows:

1) A general image restoration method termed MDDA-former is designed, which can fully exploit multi-scale structural differences of U-Net architecture by plugging CNN-based modules into the encoder-decoder and transformer blocks into the latent layer to balance between performance and efficiency.

2) The Multi-dimensional Dynamic Attention Block (MDAB) is meticulously designed to learn three complimentary attentions for convolutional kernels at a price of acceptable computational complexity, which can extract rich local contextual cues of diverse degradation information effectively.

3) An efficient and effective transformer block is proposed to capture global contextual information at the latent layer which consumes a relatively smaller number of model parameters and FLOPs while not compromising on global modeling ability.

4) Substantial experiments demonstrate that the proposed method achieves a better trade-off between performance and complexity compared with most representative SOTA methods for five image restoration tasks on 18 benchmark datasets, and also obtains superior performance for high-level vision tasks.


\section{Related Work}

\subsection{Image Restoration}

Image restoration is a long-standing low-level task, in which clean images are restored from degraded ones with noise, rain streak, or blur, etc. The image restoration methods can be categorized as two classes. One is customized image restoration methods, including image deraining \cite{Rain100, PL_rain}, {dehazing \cite{DehazeNet,AECR-Net,Zeng_PR}}, denoising \cite{CBDNet, FFDNet, ESWA}, deblurring \cite{DeblurGAN, MIMO-UNet+}, and super-resolution \cite{PL_CVPR, PL_SR}, which are designed for specific degradation. For example, rain streak binary map and rain accumulation map are utilized to indicate rain streak location and appearance prior, respectively \cite{Rain100}. 
Another line of research focuses on general image restoration\cite{MPRNet, SwinIR,MAXIM,Restormer,Uformer, PL_IR}, aiming to handle multiple types of image degradation within a single framework. Representative methods include MPRNet \cite{MPRNet}, SwinIR \cite{SwinIR}, and MAXIM \cite{MAXIM}, which exemplify recent advances based on CNNs, Transformers, and MLPs, respectively. To address the high computational cost of standard Transformers, several efficient architectures have been proposed, such as Uformer \cite{Uformer} and KiT \cite{KiT}, which adopt localized window-based self-attention mechanisms. Restormer \cite{Restormer} further improves efficiency through transposed attention and gated-dconv feed-forward network.

More recently, unified image restoration models \cite{AirNet, PromptIR, Gridformer, PerceiveIR, UniUIR} capable of addressing multiple degradations (e.g., noise, blur, rain, haze) with a single network have become increasingly popular. For example, GridFormer \cite{Gridformer} improves performance in adverse weather by combining residual dense connections with a grid-structured Transformer, facilitating hierarchical feature learning and long-range dependency modeling.

Most of the aforementioned IR methods adopt the multi-scale U-shape architecture for its inherent multi-scale feature extraction of low- and high-level features, achieving good performance. 
{Nevertheless, these methods could hardly strike a delicate balance between achieving high performance and maintaining efficient inference.}

\begin{figure}[!htbp]
	\centerline{\includegraphics[page=1,trim=0mm 0mm 0mm 0mm,width=1\linewidth]{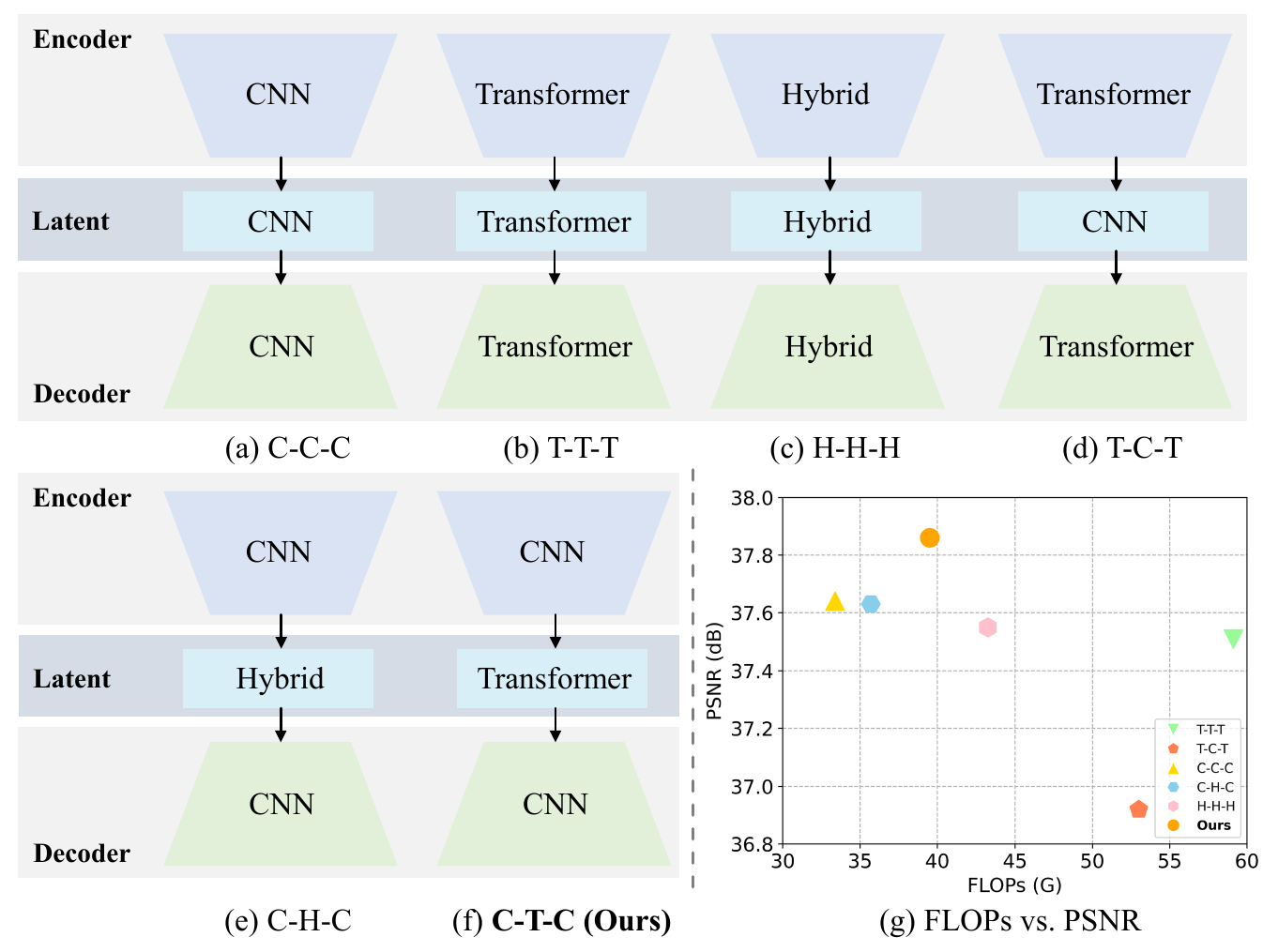}}
	\vspace{-1em}
\caption{{U-shaped image restoration architectures. C, T, and H denote CNN, Transformer, and Hybrid block (CNN + Transformer) respectively.  (a$ \sim $e) The common U-shaped image Restoration architectures. (f) Our MDDA-former adopts CNN in encoder-decoder and embeds Transformer into latent to balance the complexity and the performance. (g) The performance/FLOPs comparisons among our method and the other architectures within the same framework, as detailed in Tab. \ref{table:ablation_architecture}.}}
\label{fig:Utype}
\vspace{-1.5em}
\end{figure}

\subsection{Vision Transformer and Hybird Network}
Despite of the noticeable merits of vision transformer (ViT) \cite{ViT} in global modelling on high-level vision tasks, the computational complexity of SA \cite{SA} quadratically increases with the token size. To mitigate this issue, some ViT-based works modify SA to achieve the long-range modelling abilities with the approximately linear complexities, such as Swin Transformer with local-window SA \cite{Swin}, and XCiT with a transposed version of self-attention between keys and queries \cite{XCT}. Similar works design local-window-based Transformer blocks for low-level vision tasks, such as KiT \cite{KiT} and Uformer \cite{Uformer}. However, these transformer-based works \cite{Swin,Uformer,KiT} may sacrifice the global modelling ability due to local-window SA. In addition, pure transformers lack inductive bias of the locality which is the strength of CNNs. It is challenging to balance the complexity and the global and local modelling ability of the network.

Hybrid networks with CNNs and Transformers are initially explored in high-level vision tasks \cite{Coat,Conformer,MixFormer,Fusion}. To seek a good trade-off between model performance and efficiency, a problem of how to optimally assemble convolution and transformer layers was proposed in \cite{Coat}, and a solution of serially stacking convolutional layers first and transformer subsequently is explored. Block of parallel/dual combination of CNN and transformers with interactive information fusion were proposed in \cite{Conformer,MixFormer}. To solve the contribution assignment of CNN and transformers in local and global modeling, ASF-former mainly consisting of dual branches of CNN and transformers with an adaptive fusion scheme was proposed in \cite{Fusion}. Recently, hybrid transformers have been under investigated in low-level tasks. In \cite{SCUNet}, a Swin-Conv-UNet (SCUNet) is proposed for image denoising, in which a Swin-Conv block is designed by the combination of swin transformer and residual CNN. Asymmetric flow division unit is designed to allocate different contributions of CNN and Transformer for the hybrid CNN-Transformer block of the U-shaped architecture \cite{AFD-former}. Concurrently, it was claimed that only hybrid transformers used in the latent layer and CNNs adopted in multi-scale encoder-decoder was effective to exploit both the global and local modelling ability in \cite{Dual-former}. {Depending on the types of networks used for the encoder, decoder, and latent layer, U-shaped image restoration architectures can be approximately categorized into several groups, as illustrated in Fig. \ref{fig:Utype}.}

Similar to previous IR methods \cite{SCUNet, AFD-former, Dual-former}, we propose a general image restoration method termed MDDA-former, which aims to exploit how to unite transformer and convolution layers to exert their full potential within U-shaped architecture. The difference lies in that transformer-based and CNN-based blocks attend to their own duty by being placed in the encoder-decoder and latent layers respectively, as shown in Fig. \ref{fig:Utype} (f), and it is no bother to consider the fusion weights or interactive mechanism between them.

\vspace{-1.3em} 
\subsection{Attention Mechanism for Convolution Kernels}
The attention approaches in CNNs can be commonly categorized as feature recalibration and dynamic convolution. {Feature recalibration means calibrating output feature maps by learning and reallocating weights along channel, spatial, or multiple dimensions of these feature maps, such as SENet \cite{SE}, CBAM \cite{CBAM}, and ECA \cite{ECA}. By analogy, dynamic convolution denotes calibrating a group of convolutional kernels by learning and reallocating weights along certain dimensions of these convolutional kernels.} 
Initially, CondConv \cite{CondConv} and DyConv \cite{DyConv} employ dynamic convolution to exploit only the {kernel-wise} attention of convolutional kernels, which is restricted in capturing rich contextual cues. ODConv \cite{ODConv} designs the complementary attentions of convolutional kernels for high-level tasks to excavate four-dimensional dynamic properties. However, four-dimensional dynamic attentions, especially kernel-wise attention, will possibly result in huge computational burden.

To make a good trade-off between performance and complexity, we introduce a multi-dimensional dynamic convolution (MDConv) with spatial-wise, channel-wise and filter-wise attentions to exploit complementary attentions of convolutional kernels, which is the core component of the designed multi-dimensional dynamic attention block (MDAB). Thus, the MDAB can potentially extract rich contextual cues of various degradation information.

\begin{figure*}[!htbp]
	\centerline{\includegraphics[page=1,trim=0mm 0mm 0mm 0mm,width=1\linewidth]{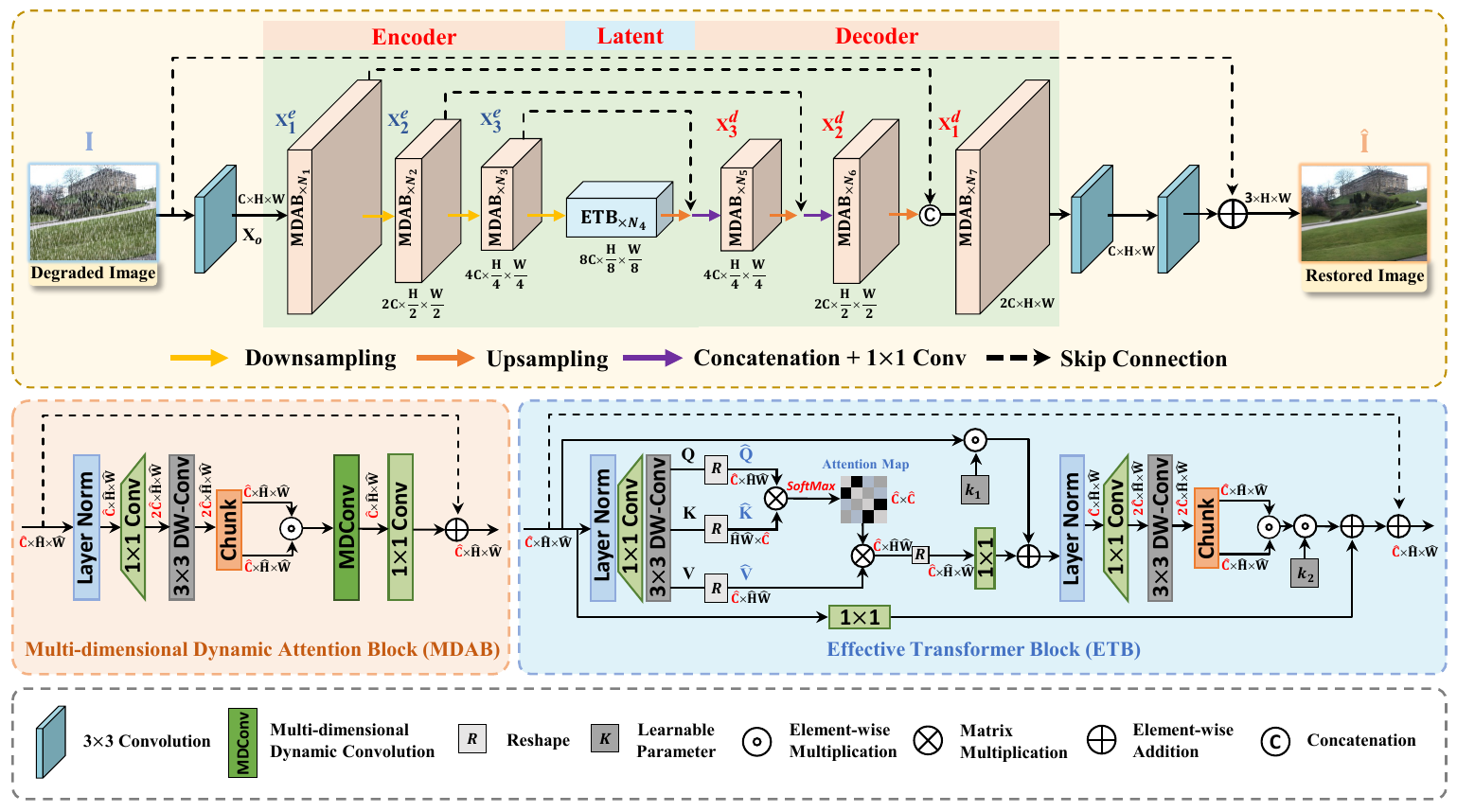}}
	\vspace{-1em}
\caption{The top part is the proposed framework of MDDA-former. For the encoder-decoder, it consists of CNN-based Multi-dimensional Dynamic Attention Block (MDAB). Effective Transformer Block (ETB) is embedded into the latent layer. The middle part is a detailed description of MDAB and ETB.}
\label{fig:framework}
\vspace{-1.5em}
\end{figure*}

\vspace{-1em} 
\section{Methodology}

\subsection{Overall Pipeline}
Fig. \ref{fig:framework} illustrates the overall architecture of the proposed general image restoration network, which is termed as MDDA-former in this paper. The MDDA-former embeds the CNN-based MDABs into the encoder-decoder. The pure transformer structure named Effective Transformer Block (ETB) is employed in the latent layer to efficiently model global semantical information, which can possibly capture the global degradation cues.

Given a degraded image $\mathbf{{I}} \in  \mathbb{R}^{\mathrm{3}\times \mathrm{H}\times \mathrm{W}}$, the MDDA-former first applies a $3\times3$ convolution with ReLU to extract feature embeddings $\mathbf{{X}}_{{o}} \in  \mathbb{R}^{\mathrm{C}\times \mathrm{H}\times \mathrm{W}}$. Then, these feature maps $\mathbf{{X}}_{{o}}$ pass through three encoders, each of which contains a stack of several MDABs and a downsampling layer. In the downsampling layer, a $3\times3$ convolution and a pixel unshuffle operation are used to double the channels
and reduce image size. Given the input feature $\mathbf{{X}}_{{o}}$, The $ i $-th encoder produces the feature maps $\mathbf{{X}}_{i} \in  \mathbb{R}^{2^{i}\mathrm{C} \times \frac {\mathrm{H}}{2^{i}} \times \frac {\mathrm{W}}{2^{i}}}$.

Unlike the previous U-shaped architectures that utilize the Transformer or CNN-Transformer blocks in the overall network \cite{Uformer,SCUNet}, we only use ETB in the latent layer to efficiently model global semantical information and to make a trade-off of efficiency and performance. In the ETB, transposed self-attention \cite{Restormer} is introduced to calculate the attention maps across the channel dimension with linear complexity. Then, the $1\times1$ point-wise and $3\times3$ depth-wise convolution with chunk operation \cite{NAF} to enrich the local contexts and to enable multiplex non-linear capability through the ETB. For the decoder, the feature maps from the latent layer pass through an upsampling layer, which is composed of a $3\times3$ convolution and a pixel shuffle operation. Then, the upsampled feature maps are concatenated with the feature maps from the last encoder via a skip connection. Next, the concatenated feature maps subsequently pass through a $1\times1$ convolution and several MDABs. Similar operations are performed in the second and third decoders. Thus, the process of decoder can be defined by
\begin{equation}
	\mathbf{X}_{i}^{d}= f_{MDAB}^{d,i}\bigg( f_{1\times1}\Big({Concat}\big(\mathbf{X}_{i+1}^{d}\uparrow,\mathbf{\hat{X}}_{i}^{e} \big)  \Big)\bigg)       ,
\end{equation}
where $ i = 1,2,3 $ and $ d, e $ represent the decoder and encoder respectively. $\uparrow$ is the up-sampling operation. $ \mathbf{\hat{X}}_{i}^{e} $ and $ \mathbf{{X}}_{i}^{d} $ denote the output features and input features of the $ i $-th encoder and decoder, respectively. $ f_{MDAB}^{d,i} $ and $ f_{1\times1} $ represent $ i $-th MDAB of the decoder and $1\times1$ convolution, respectively.  When $ i = 3 $, $ \mathbf{\hat{X}}_{i+1}^{d} $ represents the output features of the ETB. Follow \cite{Restormer}, we use MDAB to directly aggregate context information without using $ 1\times1 $ convolution to reduce the channel at level-1 decoder. Note that each stage in the U-shaped architecture contains multiple MDABs or ETBs. Finally, the restored image $ \mathbf{\hat{I}} $ is formulated as
\begin{equation}
	\mathbf{\hat{I}}= \mathbf{{I}} + f_{3\times3}\Big(f_{3\times3}\big(\mathbf{\hat{X}}_{1}^{D}\big)\Big)  .
\end{equation}

\subsection{Multi-dimensional Dynamic Attention Block}
\begin{table}[!htbp]
	\caption{{Classification of attention types and properties of several dynamic convolutions.}}
	\label{table:att_class}
		\vspace{-1em}
	\begin{center}
		\renewcommand\arraystretch{1}
		\begin{threeparttable}
			
			{\begin{tabular}{l|cccc}
			
					\toprule[1pt]
					
					\multicolumn{1}{l|}{\bf{Type}}
                        & \multicolumn{1}{c}{{Regular}}
					& \multicolumn{1}{c}{\makecell{CondConv/\\\ DyConv}}
					& \multicolumn{1}{c}{{ODConv}}
					& \multicolumn{1}{c}{\bf{MDConv}}
					\\
									
					\hline
					{Spatial}
					&\textcolor{red}{\ding{55}}  &\textcolor{red}{\ding{55}}        & \textcolor{green}{\ding{51}}   & \textcolor{green}{\ding{51}}
					\\

					{{Channel}}
					&\textcolor{red}{\ding{55}} &\textcolor{red}{\ding{55}}      &\textcolor{green}{\ding{51}}    &\textcolor{green}{\ding{51}}
					\\

                        Filter
					&\textcolor{red}{\ding{55}}   &\textcolor{red}{\ding{55}}       &\textcolor{green}{\ding{51}}    & \textcolor{green}{\ding{51}}
					\\

                        Kernel
					&\textcolor{red}{\ding{55}}     
                         &\textcolor{green}{\ding{51}}    
                         &\textcolor{green}{\ding{51}}   
                         &\textcolor{red}{\ding{55}}

					\\		
				
					\cdashline{1-5}
                        Light-weight
                        & \textcolor{green}{\ding{51}}    &\textcolor{green}{\ding{51}}    & \textcolor{red}{\ding{55}}  &\textcolor{green}{\ding{51}}
                        
                        \\
                        \bottomrule[1pt]
			\end{tabular}}
		\end{threeparttable}
	\end{center}
\vspace{-1.8em}
\end{table}

\begin{figure}[!htp]
	\centerline{\includegraphics[page=1,trim = 0mm 0mm 0mm 0mm, clip, width=1\linewidth]{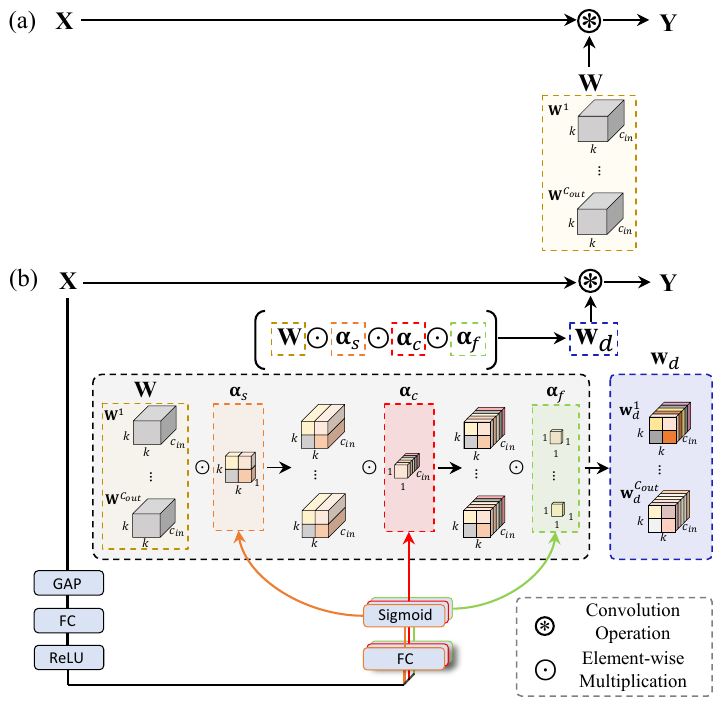}}
	 \vspace{-1em} 
	\caption{Comparison of conventional  convolution and MDConv. (a) schematic of conventional  convolution, $\mathbf{{W}}$ denotes static convolutional kernels of a group of filters, (b) schematic of MDConv, $\mathbf{{W}}_d$ represents the adjusted weight $\mathbf{{W}}$ by multiplying the learned $\bm{\alpha}_{s}$, $\bm{\alpha}_{c}$, and $\bm{\alpha}_{f}$ along spatial-wise, channel-wise, and filter-wise dimensions of $\mathbf{{W}}$.}
	\label{fig:MDConv}
	 \vspace{-1.5em}
\end{figure}

{Recently, dynamic {convolution} has been proposed to increase the capacity of a neural network without introducing much extra computational complexity \cite{CondConv,DyConv}. For image restoration, it could help enhance the network's ability to recover rich structures of the restored image. However, they only exploit one dimension of the kernel space of the convolutional filters,} which may not aggregate rich local contextual information effectively. Inspired by the omni-dimensional dynamic {convolution} \cite{ODConv}, we exploit three complementary attentions to make a trade-off between performance and complexity. These attentions include spatial-wise, channel-wise, and filter-wise attentions, which are grouped as the introduced Multi-dimensional Dynamic Convolution (MDConv). The differences among the representative dynamic {convolution} and the proposed MDConv are shown in 
Tab. \ref{table:att_class}.  Fig. \ref{fig:MDConv}(b) further demonstrates the structure of MDConv, which can be described as  
\begin{align}
        \label{eq:atten}
        & {\mathbf{Y} =  \mathbf{W}_d \circledast \mathbf{X}	,} \\ 
        & {\mathbf{W}_d = \mathbf W\odot\bm{\alpha}_{s}\odot \bm{\alpha}_{c}\odot \bm{\alpha}_{f}  ,} \\
        & {\bm{\alpha}_{s},\bm{\alpha}_{c},\bm{\alpha}_{f} = \pi\left(\mathbf{X}\right),}  
\end{align}
where $ \mathbf{X} \in \mathbb{R}^{h \times w \times C_{in}} $ is the input of the MDConv; $ \mathbf{Y} \in \mathbb{R}^{h \times w \times C_{out}} $ is the output of the MDConv; {$\mathbf W$ denotes the conventional (static) convolutional kernels for a group of $C_{out}$ filters $\mathbf{W}^m \in \mathbb{R}^{k \times k \times C_{in}}, m=1,...,C_{out} $}; ${{\bm{\alpha}_{s}}} \in \mathbb{R}^{k \times k } $, ${{\bm{\alpha}_{c}}} \in \mathbb{R}^{C_{in}} $, and ${{\bm{\alpha}_{f}}} \in \mathbb{R}^{C_{out}} $ denote three types of complementary attentions; $ \odot $ and $ \circledast $ denote element-wise multiplication and convolution operations, respectively; {$\mathbf{W}_d$ denotes the dynamic convolutional kernels adjusted from $\mathbf W$ by multiplying the learned ${\bm{\alpha}_{s}}$, ${\bm{\alpha}_{c}}$, and ${\bm{\alpha}_{f}}$  along spatial-wise, channel-wise, and filter-wise dimensions of $\mathbf{W}$;}

Following \cite{ODConv, DyConv}, $\bm{\alpha}_{s}$, $\bm{\alpha}_{c}$, and $\bm{\alpha}_{f}$ are generated by three-head attention function $\pi\left(\mathbf{X}\right)  $, which consists of a Global Average Pooling (GAP) operation, a Fully Connected (FC) layer with ReLU activation function, and three head branches (FC and Sigmoid function). {Specifically, the GAP operation is used to compress the input $ \mathbf{X}$ into a feature vector with a length of $C_{in}$. The subsequent FC layer maps the squeezed feature vector into the one with a reduced-dimensional space, characterized by a reduction ratio $r$ set to 1/4. For each of the three head branches, the FC layer outputs feature vectors/matrices with sizes of ${k \times k} $, ${C_{in}}$, and ${C_{out}} $, respectively, followed by a Sigmoid function to derive the normalized attention factors ${\bm{\alpha}_{s}}$, ${\bm{\alpha}_{c}}$, and ${\bm{\alpha}_{f}}$.}

In particular, when three complementary attentions are not used, Eq. \ref{eq:atten} can be reduced as
\begin{align}
        & {\mathbf{Y} = \mathbf W \circledast \mathbf{X} }	,
\end{align}
which is the mathematical expression for conventional convolution demonstrated in Fig. \ref{fig:MDConv}(a). 

The MDConv is inserted as a dynamic convolution into the MDAB which is embedded into the multi-scale encoder-decoder to extract rich local contextual cues of various degradation information. The structure of the MDAB is illustrated in Fig. \ref{fig:framework}. In the MDAB, a $ 1\times1 $ point-wise and $ 3\times3 $ depth-wise convolution are first applied to expand the number of dimensions and build local contexts. Then, the chunk operation \cite{NAF} is used to replace the commonly used ReLU and GELU activation functions and retain non-linear ability. Finally, the MDConv aggregates complementary multi-dimensional features. In all, the process of MDAB can be defined as  
\begin{align}
	& \mathbf{X}_{c}^{1}, \mathbf{X}_{c}^{2} =  Chunk\bigg( f_{3\times3}^{dw}\Big(  f_{1\times1} \big(  LN ( \mathbf{X}_{m})\big)  \Big) \bigg)  ,   \\
	& \mathbf{\hat{X}}_{m} = \mathbf{X}_{m} + f_{1\times1}\Big({f}_{MDconv}\big(\mathbf{X}_{c}^{1}\odot\mathbf{X}_{c}^{2}\big)  \Big)  	,
\end{align}
where $\mathbf{X}_{m}$ and $\mathbf{\hat{X}}_{m}$  represent  input and output of MDAB, respectively. {$\mathbf{X}_{c}^{1}$ and $\mathbf{X}_{c}^{2}$ denote two outputs of the chunk operation.}

\vspace{-1.5em}
\subsection{Effective Transformer Block}

Vision Transformers tend to learn global information, especially at higher layers, and it is not effective to attend to local information at early stages \cite{Like}.
In view of this, we only deploy Effective Transformer Block (ETB) into the latent layer, rather than the whole U-shaped architecture to fully exploit the ability of extracting global contextual information in the high-dimensional feature space and to make a trade-off between efficiency and performance.

As shown in Fig. \ref{fig:framework}, similar to \cite{Restormer}, the input features $ \mathbf{X}_{e} \in\mathbb{R}^{\mathrm{C}\times \mathrm{H}\times \mathrm{W}}$ are fed into Layer Norm, $ 1\times1 $ convolution, and $ 3\times3 $ depth-wise convolution to encode local contexts and generate the query $ \mathbf{Q} $, key $ \mathbf{K} $, and value $ \mathbf{V} $ projections. Then, $ \mathbf{\hat Q} $ and $ \mathbf{\hat K} $ perform matrix multiplication and pass through the Softmax function to get the attention map. The attention map $ \mathbf{\hat V} $ is matrix multiplied and then transposed after $ 1\times1 $ convolution and then summed with $ \mathbf{X}_{e} $ calibrated with learnable parameters $\mathbf{k}_{1}$ to obtain $ \mathbf{\hat{F}}_{TSA} $. This process can be described as
\begin{align}
	& \mathbf{Q, K, V}=   f_{3\times3}^{dw}\Big( f_{1\times1}\big({LN}  \left( \mathbf{X}_{e}\right) \big)\Big) , \\
        & \mathbf{\hat{Q}, \hat{K}, \hat{V}}=   \mathcal{R}\big( \mathbf{Q, K, V}\big) , 	  \\
	& \mathbf{F}_{TSA} =  SoftMax\big(\mathbf{\hat{K}}\,\otimes\,\mathbf{\hat{Q} /\alpha}\big)\,\otimes\,\mathbf{\hat{V}},  \\
	& \mathbf{\hat{F}}_{TSA} =  f_{1\times1}\big(\mathcal{R}\left(\mathbf{F}_{TSA} \right)\big)   + \mathbf{k}_{1}\odot\mathbf{X}_{e},
\end{align}
where $\mathcal{R} \left( \cdot\right) $ and $\otimes $ denote reshape operation and matrix multiplication, respectively. $ \mathbf{\hat{Q}} \in \mathbb{R}^{\mathrm{\mathrm{\hat{C}}  \times  \hat{H}}\mathrm{\hat{W}}} $; $ \mathbf{\hat{V}} \in \mathbb{R}^{\mathrm{\mathrm{\hat{C}}  \times  \hat{H}}\mathrm{\hat{W}}} $; and $ \mathbf{\hat{K}} \in \mathbb{R}^{\mathrm{\hat{H}}\mathrm{\hat{W}}  \times \mathrm{\hat{C}}} $.
For the feed-forward network (FFN), we stack a $ 1\times1 $ convolution, a $ 3\times3 $ depth-wise convolution, and a chunk operation to enrich the useful information and enable multiplex non-linear capability through the ETB. Two learnable parameters $\mathbf{{k}}_{1}$ and $\mathbf{{k}}_{2}$ for adjusting channel weights are utilized in the shortcut of the transposed self-attention and the main path of the FFN, respectively, which are initialized with identity matrices. Such design enables more flexible learning and more stable training of ETB in the latent layer.

The whole ETB process can be described as
\begin{equation}
{{\mathbf{X}_{c}^{1}, \mathbf{X}_{c}^{2}}=   {Chunk}\bigg( f_{3\times3}^{dw}\Big( f_{1\times1}\big( {LN} (
	\mathbf{\hat{F}}_{TSA}
)\big) \Big)\bigg)   ,}	  
\end{equation}
\begin{equation}
{\mathbf{\hat{X}}_{e}=\mathbf{k}_{2}\odot{\left(\mathbf{X}_{c}^{1}\odot\mathbf{X}_{c}^{2}\right)} + f_{1\times1}\left(\mathbf{X}_{e}\right) + \mathbf{X}_{e},}
\end{equation}
{where $\mathbf{{X}}_{c}^1$ and $\mathbf{{X}}_{c}^2$ represent the two outputs of the chunk operation;} $\mathbf{\hat{X}}_{e}$ denotes the output of ETB. After the ETB process, $\mathbf{\hat{X}}_{e}$ are fed into the decoder to recover local texture details layer by layer.

	 \vspace{-0.5em}
\section{Experiments and Analysis}

\begin{table}[!htbp]
\setlength{\abovecaptionskip}{2pt}
\vspace{-1.5em}
	\caption{Dataset summary on five image restoration tasks.}
	\label{table:Dataset}
	\begin{center}
 \vspace{-1em}
		\renewcommand\arraystretch{1}
		\begin{threeparttable}
			
			\resizebox{\linewidth}{!}{\begin{tabular}{l|l||c|c|c}
			
					\toprule[1pt]
                        \toprule[0.5pt]
					
					\multicolumn{1}{l|}{\bf{Task}}
					& \multicolumn{1}{l||}{\bf{Dataset}}
					& \multicolumn{1}{c|}{\bf{\#Train}}
					& \multicolumn{1}{c|}{\bf{\#Test}}
					& \multicolumn{1}{c}{\bf{Rename}}
					\\
									
					\hline
					\multirow{8}{*}{{Deraining}}
					& Rain14000 \cite{Test2800} &11200 &2800 & Test2800
					\\
					& Rain1800 \cite{Rain100} &1800 & 0 &-
					\\
					& Rain800 \cite{Test100} &700 & 100 &Test100
					\\
					& Rain12 \cite{Rain12} &12 & 0 &-
					\\
					& Rain100H \cite{Rain100} &0 & 100 &-
					\\
					& Rain100L \cite{Rain100} &0 & 100 &-
					\\
					& Rain1200 \cite{Test1200} &0 & 1200 &Test1200
					\\
	                & Raindrop \cite{AttentiveGAN} &861 & 58/239 &-
					\\
					
					\hline
					\multirow{4}{*}{{Deblurring}}
					& GoPro \cite{GoPro} &2103 &1111 &-
					\\
					& HIDE \cite{HIDE} &0 &2025 &-
					\\
					& RealBlur-J \cite{RealBlur} &0 &980 &-
					\\
					& RealBlur-R \cite{RealBlur} &0 &980 &-
					\\		
					
					\hline
					\multirow{2}{*}{{Denoising}}
					& SIDD \cite{SIDD} &320 &40 &-
					\\
					& DND \cite{DND} &0 &50 &-
					\\
					
					\hline
					\multirow{6}{*}{{Dehazing}}
					& RESIDE-Standard \cite{RESIDE} &13990 &0 &ITS
					\\
                        & SOTS-Indoor \cite{RESIDE} &0 &500 &-
					\\
					& RESIDE-V0 \cite{RESIDE} &313950 &0 &OTS
					\\
                        & SOTS-Outdoor \cite{RESIDE} &0 &500 &-
					\\
					& RESIDE-6K \cite{SOTS-Mix} &6000 &1000 &SOTS-Mix
					\\
					& Haze4K \cite{Hazk4K} &3000 &1000 &-
					\\
					
					\hline
					\multirow{2}{*}{{Enhancement}}
					& LOL-v1 \cite{Retinex-Net} &485 &15 &-
					\\
					& LOL-v2-real \cite{SGM} &689 &100 &-
					\\
     
                        \hline
					  {\multirow{6}{*}{Multi-task}}
                        &  {BSD400\cite{BSD400}} &{400} & {0} &{-}\\
                        &  {WED\cite{WED}} & {4744} &{0} &\textcolor{red}{-}\\
                        &  {CBSD68\cite{CBSD68}} &{0} &{68} &{-}\\
                        &  {RESIDE-$\beta$-OTS\cite{RESIDE}} &{72135} & {0} &\\ 
                        &  {SOTS-Outdoor\cite{RESIDE}} &{0} &{500} &{-}\\
                        &  {Rain100L\cite{Rain100}} &{200} &{100} &{-}
                        \\
					\bottomrule[1pt]
                        
			\end{tabular}}
		\end{threeparttable}
	\end{center}
\vspace{-1.5em}
\end{table}

\subsection{Experiment Settings} 

  \textbf{Training Details.} In all experiments, we trained individual MDDA-former models with PSNR loss \cite{HINet} for different image restoration tasks. The AdamW \cite{AdamW} optimizer ($\beta_1 = 0.9$, $\beta_2 = 0.999$, weight decay $ = 0.02 $) was utilized for optimizing the models. The patch size of the training image was set to 256$ \times $256, and the mini-batch was set to 16. The MDDA-former was trained on 2 GeForce RTX 3090 GPUs, with an initial learning rate of $2\textit{e} - 4$, which was decreased to $1\textit{e} - 6$ using the cosine annealing \cite{AdamW}. To augment the training data, horizontal and vertical flips were used. Note that floating point operations (FLOPs) and latency were measured on 256$ \times $256 RGB images using a single GeForce RTX 3090 GPU.

   \textbf{Implementation Details.} The feature dimensions at each stage in MDDA-former were set as [60, 120, 240, 480, 240, 120, 60]. The number of MDABs in the encoder-decoder was set as [3, 6, 6, 6, 6, 3] in MDDA-former. The number of the ETBs in the latent layer was set to 10. The results of all compared methods could be found in previously published papers, and we assumed the results reported in these papers were the best. The benchmark datasets for five image restoration tasks used in the experiments are summarized in Tab. \ref{table:Dataset}. More details about the training and testing division for each dataset are listed. 

 
\begin{table*}[!htbp]  
\setlength{\abovecaptionskip}{0pt}
	\caption{Image deraining results on Rain13K dataset. \textcolor{green}{$\bf\diamond$} denotes task-specific method, while \textcolor{orange}{$\bf\diamond$} indicates general restoration method.}
	\label{table:Deraining}
	\vspace{-1em}
	\begin{center}
		\renewcommand\arraystretch{1.02}
		\begin{threeparttable}
		\tabcolsep=0.12cm
			\resizebox{\linewidth}{!}{\begin{tabular}{l|c||cc|cc|cc|cc|cc|cc||c}
					
					\toprule[1pt]
                        \toprule[0.5pt]
					
					\multirow{2}{*}{\bf{Method}}
                        & \multirow{2}{*}{\bf Venue}
					& \multicolumn{2}{c|}{\bf {Test100}\cite{Test100}}
					& \multicolumn{2}{c|}{\bf {Rain100H}\cite{Rain100}}
					& \multicolumn{2}{c|}{\bf {Rain100L}\cite{Rain100}}
					& \multicolumn{2}{c|}{\bf {Test2800}\cite{Test2800}}
					& \multicolumn{2}{c|}{\bf {Test1200}\cite{Test1200}}
					& \multicolumn{2}{c||}{\bf {Average}}
					& \multirow{2}{*}{\makecell{\bf FLOPs(G)$\downarrow$/\\\bf \#Params(M)$\downarrow$}}
					\\  \cline{3-14}
					
                        &
					&PSNR$\uparrow$ &SSIM$\uparrow$
					&PSNR$\uparrow$ &SSIM$\uparrow$
					&PSNR$\uparrow$ &SSIM$\uparrow$
					&PSNR$\uparrow$ &SSIM$\uparrow$
					&PSNR$\uparrow$ &SSIM$\uparrow$
					&PSNR$\uparrow$ &SSIM$\uparrow$
					\\

					\hline
					\textcolor{green}{$\bf\diamond$} DerainNet \cite{DerainNet}
					& 17'TIP
                        & 22.77 &0.810
					& 14.92 &0.592
					& 27.03 &0.884
					& 24.31 &0.861
					& 23.38 &0.835
					& 22.48 &0.796
					& 88.29/\phantom{0}0.75
					\\
					
					
					
					\textcolor{green}{$\bf\diamond$} RESCAN \cite{RESCAN}
                        & 18'ECCV
					& 25.00 &0.835
					& 26.36 &0.786
					& 29.80 &0.881
					& 31.29 &0.904
					& 30.51 &0.882
					& 28.59 &0.857
					& 32.32/\phantom{0}0.15
					\\

                	\textcolor{green}{$\bf\diamond$} UMRL \cite{UMRL}
                        & 19'CVPR
					& 24.41 &0.829
					& 26.01 &0.832
					& 29.18 &0.923
					& 29.97 &0.905
					& 30.55 &0.910
					& 28.02 &0.880
					& 16.50/\phantom{0}0.98
					\\
					
					\textcolor{green}{$\bf\diamond$} PreNet \cite{PreNet}
                        & 19'CVPR
					& 24.81 &0.851
					& 26.77 &0.858
					& 32.44 &0.950
					& 31.75 &0.916
					& 31.36 &0.911
					& 29.42 &0.897
					& 66.58/\phantom{0}0.17
					\\
					
					\textcolor{green}{$\bf\diamond$} MSPFN \cite{MSPFN}
                        & 20'CVPR
					& 27.50 &0.876
					& 28.66 &0.860
					& 32.40 &0.933
					& 32.82 &0.930
					& 32.39 &0.916
					& 30.75 &0.903
					& 708.44/21.00\phantom{0}
					\\

                        \cdashline{1-15}
     
					\textcolor{orange}{$\bf\diamond$} MPRNet \cite{MPRNet}
                        & 21'CVPR
					& 30.27 &0.897
					& 30.41 &0.890
					& 36.40 &0.965
					& 33.64 &0.938
					& 32.91 &0.916
					& 32.73 &0.921
					& 148.55/\phantom{0}3.64\phantom{0}
					\\
					
					\textcolor{orange}{$\bf\diamond$} HINet \cite{HINet}
                        & 21'CVPRW
					& 30.26   &0.905
					& {30.63} &0.893
					& {37.20} &{0.969}
					& {33.87} &0.940
					& {33.01} &0.918
					& {33.00} &0.925
					& 170.73/88.67\phantom{0}
					\\
					
					\textcolor{orange}{$\bf\diamond$} KiT \cite{KiT}
                        & 22'CVPR
					& 30.26 &0.904
					& 30.47 &{0.896}
					& 36.65 &{0.970}
					& 33.85 &{0.941}
					& 32.81 &0.918
					& 32.81 &{0.926}
					& 43.08/20.60
					\\
					
					
					\textcolor{orange}{$\bf\diamond$} MAXIM-2S \cite{MAXIM}
                        & 22'CVPR
					& {31.17} &{0.922}
					& 30.81   &{0.903}
					& {38.06} &{0.976}
					& 33.80   &{0.943}
					& 32.37   &{0.921}
					& {33.24} &{0.933}
					& 216.00/14.10\phantom{0}
					\\
					
					\textcolor{orange}{$\bf\diamond$} Restormer \cite{Restormer}
					& 22'CVPR
                        & \underline{32.00}    &\underline{0.923}
					& {31.46}    &{0.904}
					& \textbf{38.99}    &\textbf{0.978}
					& \underline{34.18} &{0.944}
					& {33.19}           &{0.926}
					& \underline{33.96}    &{0.935}
					& 140.99/26.13\phantom{0}
					\\

                        \textcolor{orange}{$\bf\diamond$} SFNet \cite{SFNet}
					& 23'ICLR
                        & {31.47}    &{0.919}
					& \underline{31.90}    &{0.903}   %
					& {38.21}    &{0.974}
					& {33.69}    &{0.937}
					& {32.55}    &{0.911}
					& {33.56}    &{0.929}
					& 125.43/13.27\phantom{0}
					\\

                        \textcolor{orange}{$\bf\diamond$} {MambaIR \cite{MambaIR}}
					& {24'ECCV}
                        & {\bf{32.11}}    &{\bf{0.924}}
					& {\textbf{32.03}}    &{\bf{0.905}}   %
					& {38.46}    &{0.976}
					& {34.15}    &{\bf{0.945}}
					& {\underline{33.53}}    &{\underline{0.930}}
					& {\bf{34.06}}    &{\underline{0.936}}
					& {229.80/26.78\phantom{0}}
					\\
					
				    \textcolor{orange}{$\bf\diamond$} \cellcolor{mycolor}\textbf{MDDA-former}
                        & \cellcolor{mycolor}-
					&\cellcolor{mycolor}{31.54} &\cellcolor{mycolor}\underline{0.923}
					&\cellcolor{mycolor}{31.49} &\cellcolor{mycolor}\textbf{0.905}
					&\cellcolor{mycolor}\underline{38.51} &\cellcolor{mycolor}\textbf{0.978}
					&\cellcolor{mycolor}\textbf{34.23}    &\cellcolor{mycolor}\textbf{0.945}
					&\cellcolor{mycolor}\textbf{33.61}    &\cellcolor{mycolor}\textbf{0.932}
					&\cellcolor{mycolor}{33.88} &\cellcolor{mycolor}\textbf{0.937}
					&\cellcolor{mycolor}67.38/25.92
					\\
					
					\bottomrule[1pt]
			\end{tabular}}
		\end{threeparttable}
	\end{center}
	\vspace{-2em}
\end{table*}

\begin{figure*}[tp]  
	\centerline{\includegraphics[page=1,trim = 0mm 0mm 0mm 0mm, clip, width=1\linewidth]{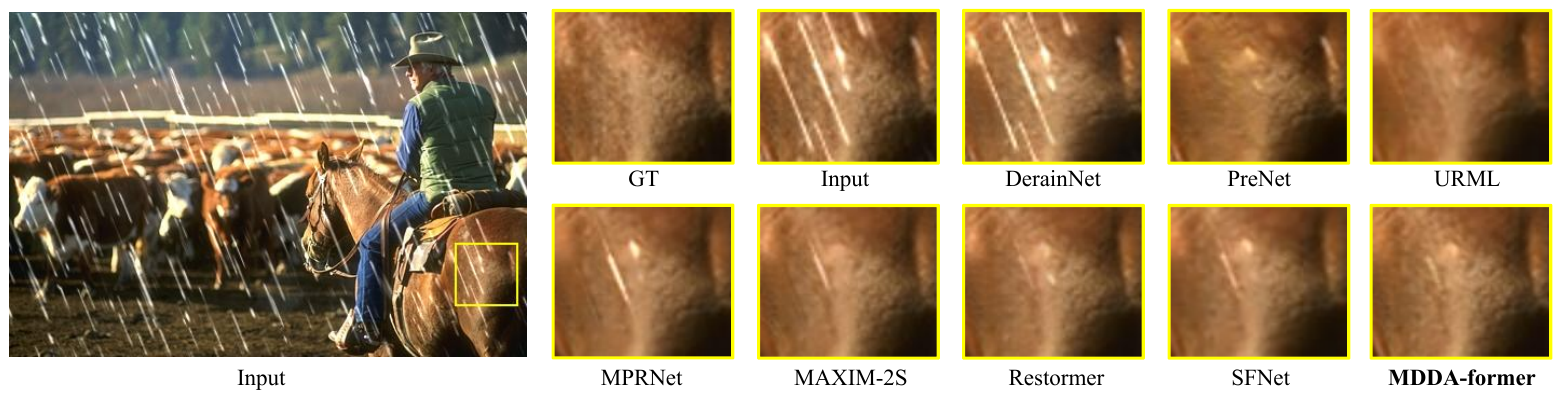}}
	\vspace{-1em}
\caption{Visual comparisons on synthetic rainy images sampled from Rain100L \cite{Rain100} dataset. \textbf{Zoom-in for best view.}}
    \label{fig:Derain}
	\vspace{-0.5em}
\end{figure*}

\begin{table}[!htbp]  
\setlength{\abovecaptionskip}{0pt}
	\caption{Raindrop removal results on Raindrop \cite{AttentiveGAN} dataset.} 
	\label{table:Raindrop}
		\vspace{-1em}
	\begin{center}
		\renewcommand\arraystretch{1.01}
		\begin{threeparttable}
			\tabcolsep=0.1cm
			\resizebox{\linewidth}{!}{\begin{tabular}{l|c||cc|cc||c}
					
					\toprule[1pt]
                        \toprule[0.5pt]
					
					\multirow{2}{*}{\bf{Method}}
                        &\multirow{2}{*}{\bf Venue}
					& \multicolumn{2}{c|}{\bf {Raindrop-A}\cite{AttentiveGAN}}
					& \multicolumn{2}{c||}{\bf {Raindrop-B}\cite{AttentiveGAN}}
					& \multirow{2}{*}{\makecell{\bf FLOPs(G)$\downarrow$/\\\bf \#Params(M)$\downarrow$}}
					\\  \cline{3-4} \cline{5-6} 
     
					&
					&PSNR$\uparrow$ &SSIM$\uparrow$
					&PSNR$\uparrow$ &SSIM$\uparrow$
					\\
									
					\hline
					pix2pix \cite{pix2pix}
                        & 17'CVPR
					& 28.02 & 0.855
					& 23.50 &0.715
					& 18.16/54.40 
					\\
					
					AttentiveGAN \cite{AttentiveGAN}
                        & 18'CVPR
					& {31.55} &0.902
					& 24.92 &0.809
					& 89.67/\phantom{0}6.24
					\\
					
					DuRN \cite{DuRN}
					& 19'CVPR
                        & 31.24 &0.926
					& 25.32 &0.817
					& 89.39/12.86
					\\
					
							
					
                        IDT\cite{IDT}
                        & 22'TPAMI
                        & 31.87 &0.931
                        & -     &-
                        & 61.90/16.00
                        \\

					MAXIM-2S\cite{MAXIM}
					& 22'CVPR
                        & 31.87 &0.935
					& \underline{25.74} &\underline{0.827}
                        & 216.00/14.10\phantom{0}
					\\
					
                        UDR-S\(^2\)Former\cite{UDR-SFormer}
					&23'CVPR
                        & \underline{32.64} &\textbf{0.943}
					& -     &-
                        & 21.58/\phantom{0}8.53
					\\
                        
                        \cellcolor{mycolor}\textbf{MDDA-former}
					& \cellcolor{mycolor}-
                        &\cellcolor{mycolor}\textbf{32.78} &\cellcolor{mycolor}\textbf{0.943}
					&\cellcolor{mycolor}\textbf{26.81} &\cellcolor{mycolor}\textbf{0.833}
					& \cellcolor{mycolor}67.38/25.92
					\\

					\bottomrule[1pt]
			\end{tabular}}
		\end{threeparttable}
	\end{center}
	\vspace{-2em}
\end{table}

\begin{figure}[tbp]  
	\centerline{\includegraphics[page=1,trim = 0mm 0mm 0mm 0mm, clip, width=1\linewidth]{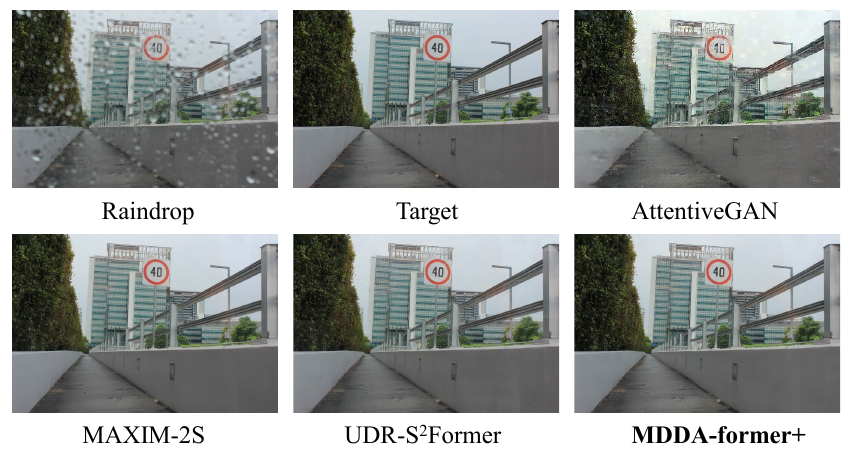}}
 \vspace{-1em}
	\caption{Visual comparisons on Raindrop \cite{AttentiveGAN} dataset.
		\textbf{Zoom-in for best view.}}
	\label{fig:Raindrop}
	\vspace{-1em}
\end{figure}  

\vspace{-1.2em} 
\subsection{Image Deraining Results}
Following the previous works \cite{MAXIM,MSPFN,KiT}, we trained the MDDA-former on 13,712 clean-rainy image pairs (Rain13K dataset) and 861 pairs Raindrop \cite{AttentiveGAN} dataset over 400K and 100K steps, respectively. The PSNR/SSIM scores are computed on the Y channel in YCbCr color space. {As shown in Tab. \ref{table:Deraining}, the proposed MDDA-former achieve the 3rd best PSNR and the best SSIM on average across the five datasets. But when considering both efficiency and effectiveness, our method could strike a good balance. Specifically, with a similar number of model parameters, MDDA-former achieves competitive performance using 28\% (67.38G vs. 229.80G) and 48\% (67.38G vs. 140.99G) FLOPs compared to MambaIR \cite{MambaIR} and Restormer \cite{Restormer}, respectively.} Although our MDDA-former has a larger number of parameters, it achieves 0.64- and 0.32 dB PSNR improvements while utilizing 54\%  (67.38G vs. 125.43G) and 32\% (67.38G vs. 216G) FLOPs compared with SFNet \cite{SFNet} and MAXIM-2S \cite{MAXIM}, respectively. As demonstrated in Tab. \ref{table:Raindrop}, the proposed MDDA-former obtains 0.91 dB and 0.14 dB gains in PSNR than IDT \cite{IDT} and UDR-S\(^2\)Former \cite{UDR-SFormer} on Raindrop-A \cite{AttentiveGAN}, respectively. Fig. \ref{fig:Derain} and Fig. \ref{fig:Raindrop} visually demonstrate that our method is effective in removing rain streaks and drops, which is consistent with the quantitative results in Tab. \ref{table:Deraining} and Tab. \ref{table:Raindrop}.

\begin{table*}[!htbp]   
\setlength{\abovecaptionskip}{0pt}
	\caption{Image deblurring results. Our method is trained on GoPro \cite{GoPro} dataset and directly evaluated on HIDE \cite{HIDE} and RealBlur \cite{RealBlur} dataset. \textcolor{green}{$\bf\diamond$} denotes task-specific method, while \textcolor{orange}{$\bf\diamond$} indicates general restoration method.}
	\label{table:Deblurring}
	\vspace{-1em}
	\begin{center}
		\renewcommand\arraystretch{1.03}
		\begin{threeparttable}
			
			\scalebox{1}{\begin{tabular}{l|c||cc|cc|cc|cc||c}
					
                        \toprule[1pt]
                        \toprule[0.5pt]
					\multirow{2}{*}{\textbf{Method}}
                        & \multirow{2}{*}{\textbf{Venue}}
					& \multicolumn{2}{c|}{\textbf{GoPro}\cite{GoPro}}
					& \multicolumn{2}{c|}{\textbf{HIDE}\cite{HIDE}}
					& \multicolumn{2}{c|}{\textbf{RealBlur-R}\cite{RealBlur}}
					& \multicolumn{2}{c||}{\textbf{RealBlur-J}\cite{RealBlur}}
					& \multirow{2}{*}{\makecell{\bf FLOPs(G)$\downarrow$/\\ \bf\#Params(M)$\downarrow$}}
					\\  \cline{3-4} \cline{5-6} \cline{7-8} \cline{9-10}

                        &
					&PSNR$\uparrow$ &SSIM$\uparrow$
					&PSNR$\uparrow$ &SSIM$\uparrow$
					&PSNR$\uparrow$ &SSIM$\uparrow$
					&PSNR$\uparrow$ &SSIM$\uparrow$
					\\
					
					\hline
                        \textcolor{green}{$\bf\diamond$} Nah \textit{et al.} \cite{GoPro}
                        & 17'CVPR
					& 29.08 &0.914
					& 25.73 &0.874
					& 32.51 &0.841
					& 27.87 &0.827
					& 336.00/11.70\phantom{0}
					\\
     
					\textcolor{green}{$\bf\diamond$} DeblurGAN \cite{DeblurGAN}
                        & 18'CVPR
					& 28.70 &0.858
					& 24.51 &0.871
					& 33.79 &0.903
					& 27.97 &0.834
					& 16.96/14.00
					\\
								
					\textcolor{green}{$\bf\diamond$} DeblurGAN-v2 \cite{DeblurGAN-v2}
                        & 19'ICCV
					& 29.55 &0.934
					& 26.61 &0.875
					& 35.26 &0.944
					& 28.70 &0.866
					& 42.00/60.93
					\\
					

                        \textcolor{green}{$\bf\diamond$} DMPHN \cite{DMPHN}
                        & 19'CVPR
					& 31.20 &0.940
					& 29.09 &0.924
					& 35.70 &0.948
					& 28.42 &0.860
					& 235.00/21.70\phantom{0}
					\\
					
					\textcolor{green}{$\bf\diamond$} DBGAN \cite{DBGAN}
                        & 20'CVPR
					& 31.10 &0.942
					& 28.94 &0.915
					& 33.78 &0.909
					& 24.93 &0.745
					& 759.85/11.59\phantom{0}
					\\
					
					\textcolor{green}{$\bf\diamond$} MTRNN \cite{MT-RNN}
                        & 20'ECCV
					& 31.15 &0.945
					& 29.15 &0.918
					& 35.79 &0.951
					& 28.44 &0.862
					& 13.72/\phantom{0}2.64
					\\
								
					
					\textcolor{green}{$\bf\diamond$} MIMO-UNet+ \cite{MIMO-UNet+}
                        & 21'ICCV
					& {32.45} &{0.957}
					& 29.99 &0.930
					& 35.54 &0.947
					& 27.63 &0.837
					& 154.41/16.11\phantom{0}
					\\

                        \textcolor{green}{$\bf\diamond$} Stripformer \cite{Stripformer}
                        & 22'ECCV
					& {33.08} &{0.962}
					& {31.03} &{0.940}
					& -       &-
					& -       &-
					& 177.43/19.71\phantom{0}
					\\

                        \textcolor{green}{$\bf\diamond$} DDANet \cite{DDANet}
                        & 23'AAAI
					& {33.07} &{0.962}
					& {30.64} &{0.937}
					& 35.81   & 0.951
					& -       &-
					& 153.51/16.18\phantom{0}
					\\
					

                        \cdashline{1-11}
                        
					\textcolor{orange}{$\bf\diamond$} MPRNet \cite{MPRNet}
                        & 21'CVPR
					& {32.66} &{0.959}
					& {30.96} &{0.939}
					& {35.99} &{0.952}
					& 28.70 &0.873
					& 777.01/20.13\phantom{0}
					\\
					

                        
                        \textcolor{orange}{$\bf\diamond$} {NAFNet} \cite{NAF}
                        & {22'ECCV}
					& {{33.08}} &{\textbf{0.963}}
					& {\textbf{31.22}} &{\textbf{0.943}}
					& {36.14} &{0.955}
					& {28.43} &{0.860}
					& {63.33}/{67.89}
					\\

     				\textcolor{orange}{$\bf\diamond$} Restormer \cite{Restormer}
                        & 22'CVPR
					& {32.92} &{0.961}
					& \textbf{31.22} &{0.942}
					& \underline{36.19} &\underline{0.957}
					& {28.96} &\underline{0.879}
					& 140.99/26.11\phantom{0}
					\\

					\textcolor{orange}{$\bf\diamond$} {MambaIR \cite{MambaIR}}
                        &{24'ECCV}
					&{\bf{33.27}} &{\bf0.963}
					&{31.11} &{0.941}
					&{36.15} &{0.955}
					&{\underline{28.99}} &{0.878}
					& {229.80}/{26.78\phantom{0}}
					\\
					
					\textcolor{orange}{$\bf\diamond$} \cellcolor{mycolor}\textbf{MDDA-former}
                        & \cellcolor{mycolor}-
					&\cellcolor{mycolor}\underline{33.21} &\cellcolor{mycolor}\textbf{0.963}
					&\cellcolor{mycolor}{31.17} &\cellcolor{mycolor}\textbf{0.943}
					&\cellcolor{mycolor}\textbf{36.26} &\cellcolor{mycolor}\textbf{0.958}
					&\cellcolor{mycolor}\textbf{29.08} &\cellcolor{mycolor}\textbf{0.881}
					& \cellcolor{mycolor}67.38/25.92
					\\
					
					\bottomrule[1pt]
			\end{tabular}}
		\end{threeparttable}
	\end{center}
	\vspace{-2em}
\end{table*}

\begin{figure*}[!htp]  
	\centerline{\includegraphics[page=1,trim = 0mm 0mm 0mm 0mm, clip, width=1\linewidth]{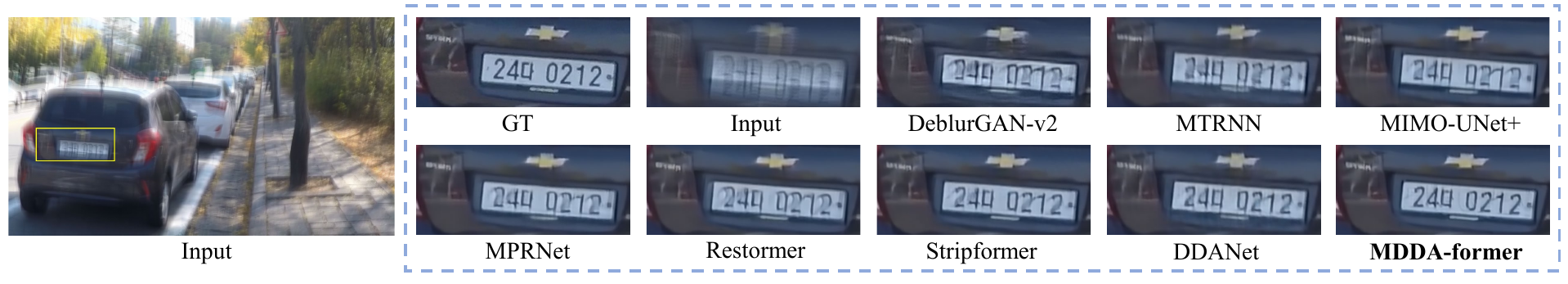}}
	\vspace{-1em}
	\caption{Visual comparisons on GoPro \cite{GoPro} dataset.
		\textbf{Zoom-in for best view.}}
	\label{fig:Deblur}
	\vspace{0em}
\end{figure*}

\begin{figure*}[!htp]  
	\centerline{\includegraphics[page=1,trim = 0mm 0mm 0mm 0mm, clip, width=1\linewidth]{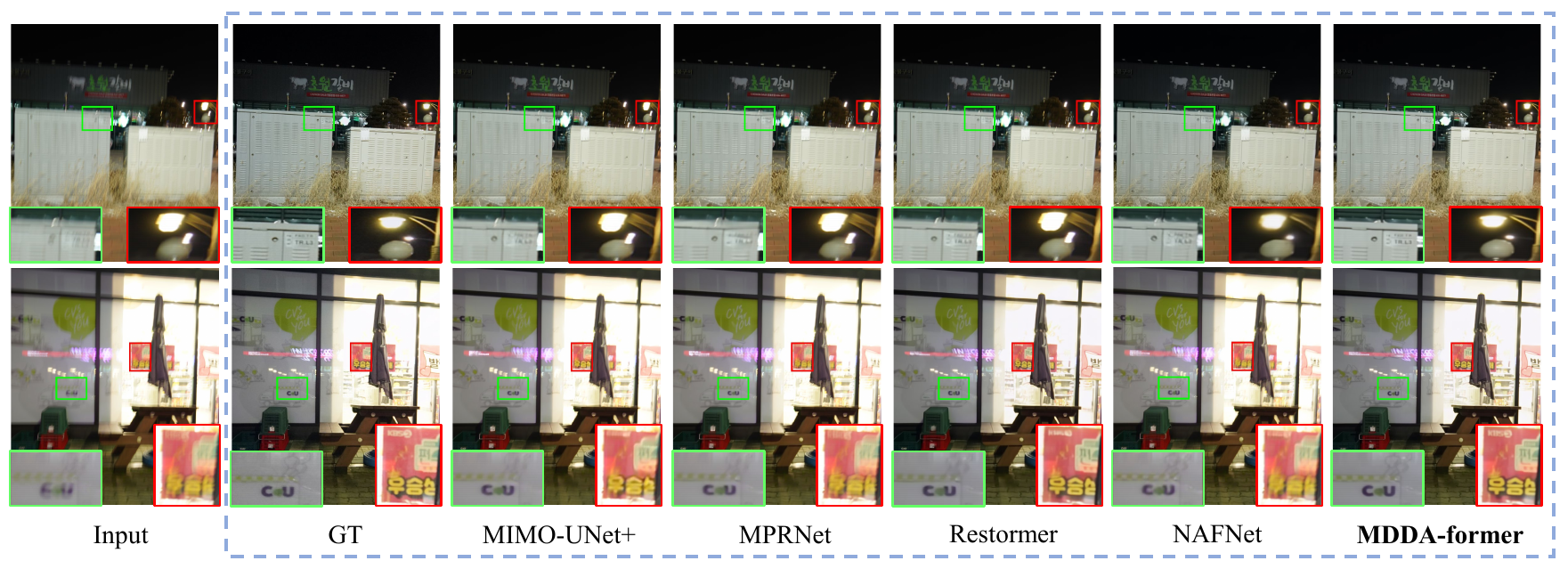}}
	\vspace{-1em}
	\caption{Visual comparisons on RealBlur-J \cite{RealBlur} dataset. \textbf{Zoom-in for best view.}}
	\label{fig:Deblur-real}
	\vspace{-2em}
\end{figure*}

\vspace{-1.2em} 
\subsection{Image Deblurring Results}
Following \cite{MPRNet,HINet}, we trained our method on GoPro \cite{GoPro} dataset with 2,103 blurry-sharp image pairs over 500K steps. Tab. \ref{table:Deblurring} demonstrates the deblurring performance comparsion. Compared with state-of-the-art (SOTA) task-specific methods (e.g., MIMO-UNet+ \cite{MIMO-UNet+}, Stripformer \cite{Stripformer}, and DDANet \cite{DDANet}), our MDDA-former achieves better performance with fewer FLOPs. Specifically, our MDDA-former outperforms MIMO-UNet+ \cite{MIMO-UNet+} on all 4 datasets using 44\% FLOPs (67.38G vs. 154.41G), surpasses Stripformer by over 0.13 and 0.14 dB on GoPro and HIDE \cite{HIDE} datasets with 38\% FLOPs (67.38G vs. 177.43G), and achieves 0.14, 0.53, and 0.45 dB PSNR gains over DDANet on GoPro, HIDE, and RealBlur-R\cite{RealBlur} datasets, respectively.
{Compared with SOTA general restoration methods (e.g., NAFNet \cite{NAF}, Restormer \cite{Restormer}, and  MambaIR \cite{MambaIR}), MDDA-former also performs better. Specifically, while achieving a lower result than Restormer and NAFNet on HIDE, it outperforms them on GoPro and RealBlur datasets. Due to the efficient Mamba \cite{Mamba} architecture, MambaIR achieved superior performance compared to other SOTA methods. While the PSNR of MDDA-former is lower than MambaIR on GoPro dataset, it outperforms MambaIR on HIDE and RealBlur datasets with less resource consumption.} {Figs. \ref{fig:Deblur} and \ref{fig:Deblur-real} demonstrate visually pleasing results on example images from the GoPro and RealBlur-J datasets, characterized by sharp edges and fine details, notably excelling in removing artifacts in the digits of the car plate and the ghosting effect of the street lamp.}

\vspace{-1em}
\begin{table*}[!htbp]  
\setlength{\abovecaptionskip}{1pt}
	\caption{Real image denoising on SIDD\cite{SIDD} and DND\cite{DND} datasets. \textcolor{green}{$\bf\diamond$} denotes task-specific method, while \textcolor{orange}{$\bf\diamond$} indicates general restoration method.}
 		\vspace{-1.0em}
	\label{table:real_noise}
	\begin{center}
\renewcommand\arraystretch{1.03}
  \tabcolsep=0.08cm
		\resizebox{\linewidth}{!}{\begin{tabular}{c|c||ccccccccccc}  
   
				\toprule[1pt]
                \toprule[0.5pt]
				
				\multirow{2}{*}{\textbf{Dataset}}
				&\multirow{2}{*}{\textbf{Evaluation}}
				& \textcolor{green}{$\bf\diamond$} {DnCNN}
				& \textcolor{green}{$\bf\diamond$} {RIDNet}
                & \textcolor{green}{$\bf\diamond$} {ADFNet}
				& \textcolor{orange}{$\bf\diamond$} {MIRNet}
				& \textcolor{orange}{$\bf\diamond$} {MPRNet}
				& \textcolor{orange}{$\bf\diamond$} {Uformer-B}
				& \textcolor{orange}{$\bf\diamond$} {MAXIM-3S}
				& \textcolor{orange}{$\bf\diamond$} {Restormer}
				& \textcolor{orange}{$\bf\diamond$} {NAFNet}
				& \textcolor{orange}{$\bf\diamond$} {{MambaIR}}
				& \multirow{2}{*}{\textcolor{orange}{$\bf\diamond$} \textbf{MDDA-former}}
				\\

                &
                & \cite{DnCNN}
				& \cite{RIDNet}
                & \cite{ADFNet}
				& \cite{MIRNet}
				& \cite{MPRNet}
				& \cite{Uformer}
				& \cite{MAXIM}
				& \cite{Restormer}
				& \cite{NAF}
				& \cite{MambaIR}
				& 
				\\
				
				\hline
				\makecell{{SIDD}\\\cite{SIDD}}
				& \makecell{PSNR$\uparrow$\\SSIM$\uparrow$}
				& \makecell{30.78 \\ 0.801}
				& \makecell{38.71 \\ 0.951}
                & \makecell{{39.63} \\ {0.958}}
				& \makecell{39.72 \\ {0.959}}
				& \makecell{39.71 \\ 0.958} 
				& \makecell{39.89 \\ \underline{0.960}}
				& \makecell{{39.96} \\ \underline{0.960}}
				& \makecell{\underline{40.02} \\ \underline{0.960}}
				& \makecell{{\textbf{40.30}} \\ {\textbf{0.962}}} 
                & \makecell{{39.89}\\{\underline{0.960}}}
				& \makecell{{39.96}\\ \underline{0.960}}
				\\
				
				\hline
				\makecell{{DND}\\\cite{DND}}
				& \makecell{PSNR$\uparrow$\\SSIM$\uparrow$}
				& \makecell{32.43 \\0.790}
				& \makecell{39.26 \\0.953}
                & \makecell{{39.89} \\ {0.960}}
				& \makecell{39.88 \\\textbf{0.956}}
				& \makecell{39.80 \\0.954} 
				& \makecell{{39.98} \\ {0.955}}
				& \makecell{{39.84} \\ {0.954}}
				& \makecell{{40.03} \\ \textbf{0.956}}
				& \makecell{{-}\\ {-}} 
                & \makecell{\underline{40.04}\\ \bf0.956}
				& \makecell{\textbf{40.05} \\ \textbf{0.956}}
				\\
				
				\hline
                & \textbf{Venue}
                & 17'TIP
                & 19'ICCV
                & 23'AAAI
                & 20'ECCV
                & 21'CVPR
                & 22'CVPR
                & 22'CVPR
                & 22'CVPR
                & {22'ECCV}  
                & {24'ECCV}
                & {-}
                \\

                & \makecell{\textbf{FLOPs (G)$\downarrow$}\\\textbf{\#Params (M)$\downarrow$}}
				& \makecell{43.94 \\0.67}
				& \makecell{98.13 \\1.50}
                & \makecell{55.64\\ 7.65}
				& \makecell{786.43 \\31.79}
				& {\makecell{148.55 \\3.64}} 
				& \makecell{89.46\\ 50.88}
				& \makecell{320.32 \\20.60}
				& \makecell{140.99 \\26.11}
				& \makecell{{63.33}\\ {67.89}}  
                & \makecell{{229.80}\\{26.78}}
				& {\makecell{67.38\\ 25.92}}
                \\
		
				\bottomrule[1pt]
		\end{tabular}}
		\vspace{-1.5em}
	\end{center}
	
\end{table*}

\begin{figure*}[!htp] 
	\vspace{0em}
	\centerline{\includegraphics[page=1,trim = 0mm 0mm 0mm 0mm, clip, width=1\linewidth]{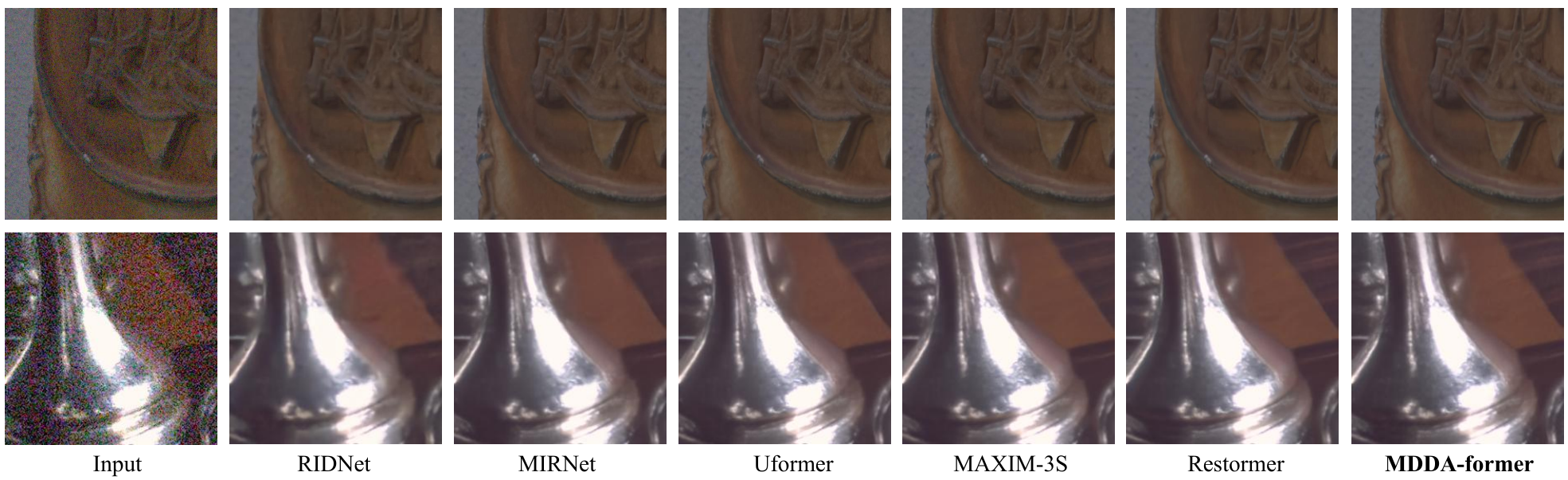}}
	\caption{Visual comparisons on real-world noisy images. Top row: image sampled from DND \cite{DND} dataset. Bottom
		row: image sampled from SIDD \cite{SIDD} dataset. \textbf{Zoom-in for best view.}}
	\label{fig:real_noise}
	\vspace{-1em}  
\end{figure*}

\subsection{Image Denoising Results}
Following \cite{Uformer,Restormer}, for real-world image denoising, the 320 images of SIDD \cite{SIDD} were used for training while the remaining 40 images from SIDD and the 1000 patches from DND \cite{DND} were used for testing.
As shown in Tab. \ref{table:real_noise}, MDDA-former obtains the best and third best PSNR results on DND and SIDD datasets, respectively. {Concretely, MDDA-former adopts 48\% (67.38G vs. 140.99G) and 21\% (67.38G vs. 229.80G) FLOPs while surpassing Restormer \cite{Restormer} and MambaIR \cite{MambaIR} over 0.02 and 0.01 dB on DND, respectively. }
{The NAFNet \cite{NAF} achieves 0.34 dB PSNR improvement than our method on the SIDD dataset. However, with having almost the same inference time (32.75ms vs. 33.43ms, as detailed in Tab. \ref{table:complexity}) and FLOPs (67.38G vs. 63.33G), our MDDA-former only costs 38\% (25.92M vs. 67.89M) parameters of NAFNet.} It's worth noting that despite MDDA-former having a greater number of model parameters and FLOPs compared with ADFNet \cite{ADFNet}, it achieves superior performance and a 1.5$\times$ faster inference speed (32.75ms vs. 48.87ms).
Fig. \ref{fig:real_noise} shows the visual denoising results on example images from both SIDD and DND datasets. It can be observed that MDDA-former could achieve good visual results which is consistent with quantitative analysis.

\begin{table*}[!htbp]  
\setlength{\abovecaptionskip}{2pt}
	\caption{Image dehazing results on the {SOTS-Indoor}\cite{RESIDE}, {SOTS-Outdoor}\cite{RESIDE}, SOTS-Mixer\cite{SOTS-Mix}, and Haze4K\cite{Hazk4K} datasets. \textcolor{green}{$\bf\diamond$} denotes task-specific method, while \textcolor{orange}{$\bf\diamond$} indicates general restoration method. ``-'' denotes the result is unavailable.} 
	\label{table:dehaze}
	\vspace{-1em}
	\begin{center}
		\renewcommand\arraystretch{1.01}
		\begin{threeparttable}
			
			\scalebox{1}{\begin{tabular}{l|c||cc|cc|cc|cc||c}
					
					\toprule[1pt]
                        \toprule[0.5pt]
					
					\multirow{2}{*}{\textbf{Method}}
                        & \multirow{2}{*}{\textbf{Venue}}
					& \multicolumn{2}{c|}{\textbf{SOTS-Indoor}\cite{RESIDE}}
					& \multicolumn{2}{c|}{\textbf{SOTS-Outdoor}\cite{RESIDE}}
					& \multicolumn{2}{c|}{\textbf{SOTS-Mix}\cite{SOTS-Mix}}
					& \multicolumn{2}{c||}{\textbf{Haze4K\cite{Hazk4K}}}
					& \multirow{2}{*}{\makecell{\textbf{FLOPs (G)$\downarrow$/}\\\textbf{\#Params (M)$\downarrow$}}}
					\\   \cline{3-4} \cline{5-6} \cline{7-8} \cline{9-10}

                        &
					&PSNR$\uparrow$ &SSIM$\uparrow$
					&PSNR$\uparrow$ &SSIM$\uparrow$
					&PSNR$\uparrow$ &SSIM$\uparrow$
					&PSNR$\uparrow$ &SSIM$\uparrow$
					\\
					
					\hline
					\textcolor{green}{$\bf\diamond$} DCP \cite{DCP}
                        & 09'CVPR
					& 16.62 &0.818
					& 19.14 &0.861
					& 17.88 &0.816
					& 14.01 &0.760
					& -
					\\
					
					\textcolor{green}{$\bf\diamond$} DehazeNet \cite{DehazeNet}
                        & 16'TIP
					& 19.82 &0.821
					& 24.75 &0.927
					& 21.02 &0.870
					& 19.12 &0.840
					& \phantom{0}0.58/0.009
					\\
					
					\textcolor{green}{$\bf\diamond$} MSCNN \cite{MSCNN}
                        & 16'ECCV
					& 19.84 &0.833
					& 22.06 &0.908
					& 20.31 &0.863
					& 14.01 &0.510
					& \phantom{0}0.53/0.008
					\\
					
					\textcolor{green}{$\bf\diamond$} AOD-Net \cite{AOD-Net}
                        & 17'ICCV
					& 20.51 &0.816
					& 24.14 &0.920
					& 20.27 &0.855
					& 17.15 &0.830
					& \phantom{0}0.12/0.002
					\\
					
					
					
					\textcolor{green}{$\bf\diamond$} GridDehazeNet \cite{GDN}
                        & 19'ICCV
					& 32.16 &0.984
					& 30.86 &0.982
					& 25.86 &0.944
					& 23.29 &0.930
					& 21.49/\phantom{0}0.96
					\\
					
					\textcolor{green}{$\bf\diamond$} MSBDN \cite{MSBDN}
                        & 20'CVPR
					& 33.67 &0.985
					& 33.48 &0.982
					& 28.56 &0.966
					& 22.99 &0.850
					& 41.54/31.35
					\\
					
					
					\textcolor{green}{$\bf\diamond$} FFA-Net \cite{FFA-Net}
                        & 20'AAAI
					& 36.39 &0.989
					& 33.57 &0.984
					& 29.96 &{0.973}
					&  {26.96} &  {0.950}
					& 287.80/\phantom{0}4.46\phantom{0}
					\\
					
					\textcolor{green}{$\bf\diamond$} AECR-Net \cite{AECR-Net}
                        & 21'CVPR
					& {37.17} &0.990
					& -     &-
					& 28.52 &0.964
					& -     &-
					& 52.20/\phantom{0}2.61
					\\
					
					\textcolor{green}{$\bf\diamond$} DeHamer \cite{DeHamer}
                        & 22'CVPR
					& 36.63 &0.988
					& {35.18} &{0.986}
					&-      &-
					& -     &-
					& 48.93/132.5
					\\

     				\textcolor{green}{$\bf\diamond$} DehazeFormer-M \cite{RS-Haze}
                        & 23'TIP
					& {38.46}           &{0.994}
					& 34.29             &0.983
					& {30.89} &{0.977}
					& -                 &-
					& 48.64/\phantom{0}4.63
					\\
     
                        \textcolor{green}{$\bf\diamond$} DehazeFormer-L \cite{RS-Haze}
                        & 23'TIP
					& {40.05}           &\textbf{0.996}
					& -                 &-
					& -                 &-
					& -                 &-
					& 279.70/25.44\phantom{0}
					\\

                        \textcolor{green}{$\bf\diamond$} MITNet \cite{MITNet}
                        & 23'ACM MM
					& {40.23}           &{0.992}
					&  {35.18}             &\textbf{0.988}
					& - &-
					& -                 &-
					& 16.42/\phantom{0}2.73
					\\

                        \cdashline{1-11}
                        
					\textcolor{orange}{$\bf\diamond$} MAXIM-2S \cite{MAXIM}
                        & 22'CVPR
					& {38.11} &0.991
					& 34.19   &{0.985}
					& -       &-
					& -       &-
					& 216.00/14.10\phantom{0}
					\\
							
					

					\textcolor{orange}{$\bf\diamond$} {MambaIR \cite{MambaIR}}
                        &{24'ECCV}
					&{\underline{40.58}} &{0.995}
					&{\underline{35.41}} &{0.986}
					&{\underline{31.24}} &{\bf 0.979}
					&{\underline{33.17}} &{\underline{0.987}}
					& {229.80}/{26.78\phantom{0}}
					\\

					\textcolor{orange}{$\bf\diamond$} \cellcolor{mycolor}\textbf{MDDA-former}
                        & \cellcolor{mycolor}-
					&\cellcolor{mycolor}\textbf{40.79} &\cellcolor{mycolor}\textbf{0.996}
					& \cellcolor{mycolor}\textbf{36.27}&\cellcolor{mycolor}\textbf{0.988}
					&\cellcolor{mycolor}\textbf{31.56} &\cellcolor{mycolor}\textbf{0.979}
					&\cellcolor{mycolor}\textbf{33.43} &\cellcolor{mycolor}\textbf{0.988}
					& \cellcolor{mycolor}67.38/25.92
					\\
                        \bottomrule
                
			\end{tabular}}
		\end{threeparttable}
	\end{center}
\vspace{-1.5em}
\end{table*}

\begin{figure*}[htp]  
	\centerline{\includegraphics[page=1,trim = 0mm 0mm 0mm 0mm, clip, width=1\linewidth]{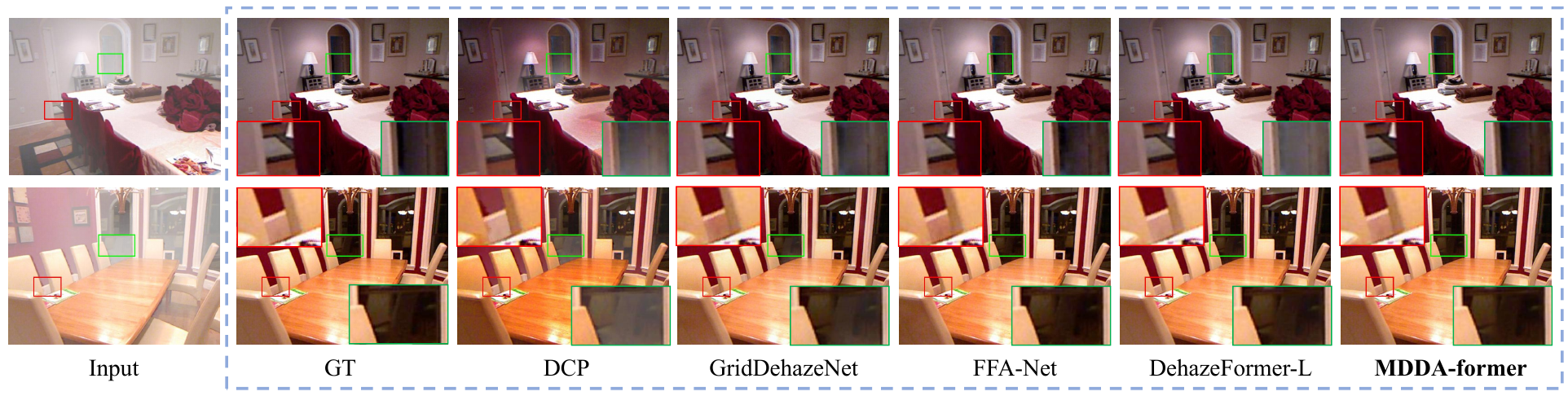}}
	\caption{Visual comparisons on a synthetic hazy image sampled from SOTS-Indoor \cite{RESIDE} dataset.
		\textbf{Zoom-in for best view.}
	}
	\vspace{-1em}
	\label{fig:Dehaze}
	\vspace{1em}
\end{figure*}

\begin{table*}[!htbp]  
\vspace{-1.5em}
\setlength{\abovecaptionskip}{1pt}
\caption{Image enhancement results on LOL \cite{Retinex-Net} and LOL-v2-real \cite{SGM} datasets. \textcolor{green}{$\bf\diamond$} denotes task-specific method, while \textcolor{orange}{$\bf\diamond$} indicates general restoration method.} 
\vspace{-1em} 
	\label{table:Low_light}
	\begin{center}
		\renewcommand\arraystretch{1.03}
		\begin{threeparttable}
			
			\scalebox{1}{\begin{tabular}{l|c||ccc|ccc||c}
					
					\toprule[1pt]
					\toprule[0.5pt]
					\multirow{2}{*}{\textbf{Method}}
                        & \multirow{2}{*}{\textbf{Venue}}
					& \multicolumn{3}{c|}{\textbf{LOL-v1} \cite{Retinex-Net}}
					& \multicolumn{3}{c||}{\textbf{LOL-v2-real} \cite{SGM}}
					& \multirow{2}{*}{\makecell{\textbf{FLOPs (G)$\downarrow$/}\\\textbf{\#Params (M)$\downarrow$}}}
					\\  \cline{3-5} \cline{6-8}

                        &
					&PSNR$\uparrow$ &SSIM$\uparrow$
					&LPIPS$\downarrow$
					&PSNR$\uparrow$ &SSIM$\uparrow$
					&LPIPS$\downarrow$
					\\
					
					\hline
					
					
					
					\textcolor{green}{$\bf\diamond$} Retinex-Net \cite{Retinex-Net}
                        & 18'BMVC
					& 16.77 &0.560
					&0.381
					
					& 18.37 &0.723
					&0.425
					
					& 587.47/\phantom{0}0.84\phantom{0}
					\\
					
					\textcolor{green}{$\bf\diamond$} KinD \cite{KinD}
                        & 19'ACM MM
					& 20.86 &0.790
					&0.170
					
					& 19.74 &0.761
					&0.362
					
					& 34.99/\phantom{0}8.16
					\\
					
					
					
					
					\textcolor{green}{$\bf\diamond$} Zero-DCE \cite{Zero-DCE}  
                        & 20'CVPR
					& 14.83 &0.531
					&0.335
					
					& 14.32 &0.511
					&0.483
					
					& \phantom{0}5.17/\phantom{0}0.08
					\\

					\textcolor{green}{$\bf\diamond$} KinD++ \cite{KinD++}
                        & 21'IJCV
					& 21.30 &0.821
					&0.160
					
					& -     &-
					&-
					
					& 743.95/\phantom{0}8.28\phantom{0}   
					\\
					
					\textcolor{green}{$\bf\diamond$} EnlightenGAN \cite{EnlightenGAN}  
                        & 21'TIP
					& 17.48 &0.650
					&0.322
					
					& 18.23 &0.617
					&0.421
					
					& 61.01/\phantom{0}8.64
					\\
					
					\textcolor{green}{$\bf\diamond$} Sparse \cite{SGM}
                        & 21'TIP
					& 17.20 &0.640
					&-
					
					& 20.06 &0.816
					&-
					
					& 53.26/\phantom{0}2.33
					\\
					
					\textcolor{green}{$\bf\diamond$} DRBN \cite{DRBN}
                        & 21'TIP
					& 19.55 &0.746
					&0.316
					
					& {20.13} &{0.820}
					&0.304
					
					& 48.61/\phantom{0}5.27  
					\\

     				\textcolor{green}{$\bf\diamond$} URetinex-Net \cite{URetinex-Net}
                        & 22'CVPR
					& 21.33 &{0.835}
					& {0.122}
					
					& -     &-
					&-
					
					& 56.93/\phantom{0}0.40
					\\

     				\textcolor{green}{$\bf\diamond$} UHDFour \cite{UHDFour}
                        & 23'ICLR
					& {23.09} &\textbf{0.870} & -
					&  {21.78}     &\textbf{0.870}   &-
                        & \phantom{0.00}\,-/17.54
					\\

                        \cdashline{1-9}
                        
%
%

					
					

                        \textcolor{orange}{$\bf\diamond$} MIRNet \cite{MIRNet}
                        & 20'ECCV
					& \underline{24.14} &0.830
					&-
					
					& 20.02    &0.820
					&-
					
					& 787.39/31.79\phantom{0} 
					\\
     
					\textcolor{orange}{$\bf\diamond$} Restormer \cite{Restormer}
                        & 22'CVPR
					& {22.43} &0.823
					&0.141
					
					& 19.94 &{0.827}
					& {0.292}
					
					& 140.99/26.13\phantom{0}
					\\
					
%
					\textcolor{orange}{$\bf\diamond$} MAXIM-2S \cite{MAXIM}
                        & 22'CVPR
					& {23.43} &{0.863} & -
					& -     &-   &-
                        & 216.00/14.10\phantom{0}
					\\

		
					

     				\textcolor{orange}{$\bf\diamond$} {MambaIR \cite{MambaIR}}
                        & {24'ECCV}
					&{24.05} &{0.857}
					&{\underline{0.113}}
		
					&{\underline{22.11}} &{0.834}
					&{\underline{0.262}}
					
					& {229.80}/{26.78\phantom{0}}
					\\
     
                        \textcolor{orange}{$\bf\diamond$} \cellcolor{mycolor}\textbf{MDDA-former}
                        & \cellcolor{mycolor}-
					&\cellcolor{mycolor}\textbf{24.27}     
                        &\cellcolor{mycolor}\underline{0.868}
					&\cellcolor{mycolor}\textbf{0.096}
		
					&\cellcolor{mycolor}\textbf{22.43} 
                        &\cellcolor{mycolor}\underline{0.839}
					&\cellcolor{mycolor}\textbf{0.233}
					
					& \cellcolor{mycolor}67.38/25.92
					\\
					
					\bottomrule[1pt]
			\end{tabular}}
		\end{threeparttable}
	\end{center}
	\vspace{-2.3em}
\end{table*}

\begin{figure*}[!htp]  
	\vspace{-1em}
	\centerline{\includegraphics[page=1,trim = 0mm 0mm 0mm 0mm, clip, width=1\linewidth]{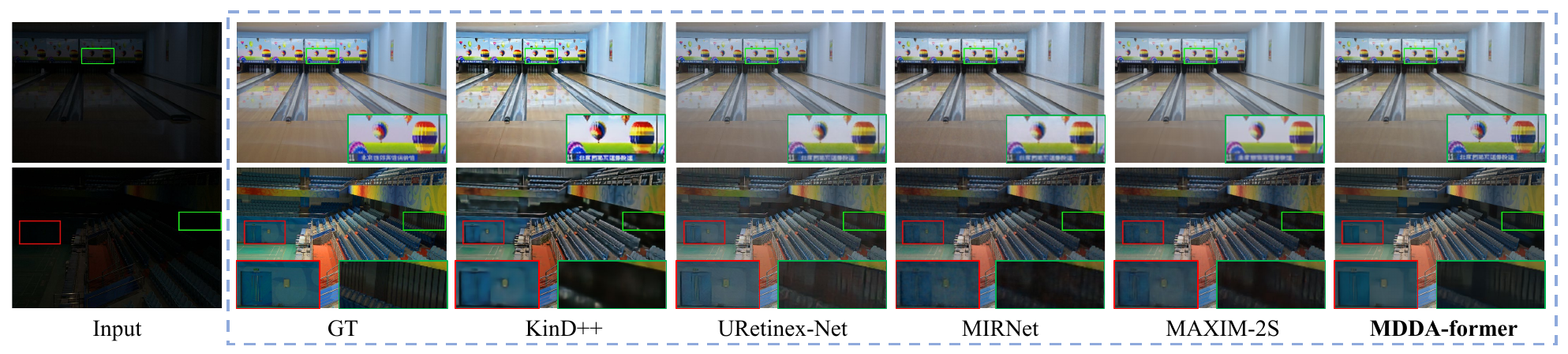}}
\vspace{-0.5em}
\caption{Visual comparison on LOL-v1 \cite{Retinex-Net} dataset. \textbf{Zoom-in for best view.}}
\label{fig:low_light}
	\vspace{-1.5em}
\end{figure*}

\vspace{-1.3em} 
\subsection{Image Dehazing Results}
The test datasets for dehazing include SOTS-Indoor \cite{RESIDE}, SOTS-Outdoor \cite{RESIDE}, SOTS-Mix \cite{SOTS-Mix}, and Haze4k \cite{Hazk4K}, covering both indoor and outdoor scenarios. SOTS-Mix dataset contains 3,000 image pairs from SOTS-Indoor and SOTS-Outdoor datasets, sourced from RESIDE \cite{RESIDE} dataset. Following DehazeFormer \cite{RS-Haze}, a few images were excluded during training since these images are smaller than the configured patch size.
As illustrated in Tab. \ref{table:dehaze}, our method achieves the best performance on all four datasets with a similar model size to DehazeFormer-L \cite{RS-Haze}, balancing performance and efficiency. Specifically, MDDA-former utilizes only 19\% of the parameters (25.92M vs. 132.45M) while outperforming DeHamer \cite{DeHamer} by 4.16 and 1.09 dB in PSNR on SOTS-Indoor and SOTS-Outdoor datasets, respectively. With comparable model parameters, MDDA-former surpasses DehazeFormer-L \cite{RS-Haze} by 0.74 dB in PSNR on SOTS-Indoor dataset, employing only 24\% of the FLOPs (67.38G vs. 279.7G). Despite having more model parameters and FLOPs than MITNet \cite{MITNet}, MDDA-former achieves 0.56 and 1.09 dB gains in PSNR and faster inference speed (32.75ms vs. 39.38ms, as detailed in Tab. \ref{table:complexity}) on SOTS-Indoor and SOTS-Outdoor datasets, respectively. {When compared with state-of-the-art general methods, MDDA-former still outperform MAXIM-2S \cite{MAXIM} and MambaIR \cite{MambaIR} by 2.68/2.08 dB and 0.21/0.86
dB in PSNR on SOTS-Indoor and SOTS-Outdoor datasets, respectively.} Fig. \ref{fig:Dehaze} shows that our proposed model can recover the hazy images with more vivid colors, e.g., the recovered table is closer to the tone of the table in the target image.

\vspace{-1.1em} 
\subsection{Low Light Image Enhancement Results}  
In the low-light image enhancement task, we used three full-reference metrics for evaluation: PSNR, SSIM, and Learned Perceptual Image Patch Similarity (LPIPS) \cite{LPIPS}. As illustrated in Tab. \ref{table:Low_light}, MDDA-former achieves the highest PSNR and LPIPS values, and ranks second in SSIM on both LOL-v1 \cite{Retinex-Net} and LOL-v2-real \cite{SGM} datasets. {Specifically, compared with state-of-the-art (SOTA) general restoration methods (e.g., MIRNet \cite{MIRNet}, Restormer \cite{Restormer}, and MambaIR \cite{MambaIR}), MDDA-former obtains gains of 0.13/1.84/0.22 dB and 2.41/2.49/0.32 dB in PSNR on LOL-v1 and LOL-v2-real datasets, respectively, with fewer FLOPs and parameters, and faster inference speed (seeing in Tab. \ref{table:complexity}).} When compared with SOTA task-specific methods (e.g., URetinex-Net \cite{URetinex-Net} and UHDFour \cite{UHDFour}), MDDA-former also demonstrates superior performance, despite having more model parameters. Fig. \ref{fig:low_light} illustrates the visual comparisons on LOL-v1 dataset. It can be clearly observed that the proposed  MDDA-former could recover more details from two low-light example images.

\vspace{-1.1em} 
\subsection{Discussions} \label{Discussions}   
\noindent\textbf{Performance and Complexity.} 
As discussed in the preceding sections, the proposed method demonstrates superior performance across five image restoration tasks. The performance and complexity comparisons on representative datasets for these tasks, involving several state-of-the-art task-agnostic and task-specific methods, are compliled and presented in Tab. \ref{table:complexity}. 
In comparison to current state-of-the-art task-agnostic methods, our MDDA-former lags behind Restormer by 0.08 dB in PSNR on Rain13K dataset. Conversely, our proposed MDDA-former outperforms Restormer on the GoPro and DND datasets, with PSNR improvements of 0.29- and 0.02 dB respectively. Notably, while having a similar number of parameters, our MDDA-former only costs 47\% FLOPs and runs 2.5$ \times $ faster than Restormer. 
Regarding the state-of-the-art task-specific methods, our MDDA-former achieves PSNR gains of 0.13- and 0.74 dB over Stripformer and Dehazeformer-L on the GoPro and LOL-v1 datasets respectively, while maintaining faster inference speed. Although the inference speed of MDDA-former is slower than that of DeHamer and URetinex-Net, it achieves substantial PSNR improvements of 4.16- and 2.94 dB on the SOTS-Indoor and LOL-v1 datasets respectively.

In conclusion, the proposed method adeptly strikes a balance between performance and complexity. This is primarily attributed to two key factors. Firstly, the well-architected model strategically deploys MDABs and ETBs in the encoder-decoder and latent layers respectively, enabling effective capture of both local and global information. Secondly, the specially designed MDAB learns complementary attentions for convolutional kernels, which can effectively extract rich local contextual cues from degraded areas.

\begin{table}[!tb]
\setlength{\abovecaptionskip}{1pt}
	\caption{Quantitative and efficiency comparisons on several representative datasets. FLOPs (G) and Latency (ms) are tested on 256$\times$256 RGB images. \textcolor{green}{$\bf\diamond$} denotes task-specific method, while \textcolor{orange}{$\bf\diamond$} indicates general restoration method.} 
	\label{table:complexity}
	\begin{center}
		\vspace{-1.8em}
		\renewcommand\arraystretch{1.1}	
  \tabcolsep=0.1cm
		\resizebox{\linewidth}{!}{\begin{tabular}{l|l|c|ccc}
				\toprule[1pt]
                \toprule[0.5pt]
				
				\textbf{Dataset}
				&\textbf{Method}
				&\textbf{PSNR}
				&\textbf{FLOPs (G)}
				&\textbf{\#Params (M)}
				&\textbf{Latency (ms)}
				\\
				
				\hline
				\multirow{5}{*}{{Rain13K}}

				&\textcolor{green}{$\bf\diamond$} MSPFN \cite{MSPFN}
				& 30.75              & 708.44  &21.00  & 53.15\\	
    

                        
				& \textcolor{orange}{$\bf\diamond$} MAXIM-2S \cite{MAXIM}
				& {33.24}            & 216.00  &14.10  & 39.27 \\
				
				&\textcolor{orange}{$\bf\diamond$} Restormer \cite{Restormer}
				& \textbf{33.96}     & 140.99  &26.13  & 81.76 \\

                 &\textcolor{orange}{$\bf\diamond$} SFNet \cite{SFNet}
				& {33.56}     & 125.43  &13.27  & 28.37 \\
				
				
				&\textcolor{orange}{$\bf\diamond$} \textbf{MDDA-former}
				& \underline{33.88}  & \phantom{0}67.38   &25.92  & 32.75 \\

				\hline
                    \multirow{4}{*}{{Raindrop}}  
                &\textcolor{green}{$\bf\diamond$} AttentiveGAN \cite{AttentiveGAN}
				& 31.55             & \phantom{0}89.67  &\phantom{0}6.24  &12.46  \\
                    
				&\textcolor{green}{$\bf\diamond$} IDT \cite{IDT}
				& 31.87             & \phantom{0}61.90  &16.00  &78.68  \\
				
				
				&\textcolor{green}{$\bf\diamond$} UDR-S$^2$Former \cite{Restormer}
				& \underline{32.64}  & \phantom{0}21.58  &\phantom{0}8.53  & 92.20 \\
				
				
				&\textcolor{orange}{$\bf\diamond$} \textbf{MDDA-former}
				& \textbf{32.78}     & \phantom{0}67.38   &25.92  & 32.75 \\

				
				
				
				
				

                    \hline
                    \multirow{7}{*}{{GoPro}}  

                    
				&\textcolor{green}{$\bf\diamond$} MIMO-UNet+ \cite{MIMO-UNet+}
				& 32.45             & 154.41  &16.11  &27.63  \\

                &\textcolor{green}{$\bf\diamond$} Stripformer \cite{Stripformer}
				& {33.08}  & 177.43 &19.71  & 39.37 \\

                &\textcolor{green}{$\bf\diamond$} DDANet \cite{DDANet}
				& {33.07}  & 153.51  &16.18  & 26.58 \\
					
				&\textcolor{orange}{$\bf\diamond$} Restormer \cite{Restormer}
				& {32.92}  & 140.99  &26.13  & 81.76 \\

                &\textcolor{orange}{$\bf\diamond$} {NAFNet \cite{NAF}}
				& {33.08}  & \phantom{0}63.33  &{67.89}  & {33.43} \\

                &\textcolor{orange}{$\bf\diamond$}  {MambaIR \cite{MambaIR}}
				&  {\bf{33.27}}  &  {229.80}   & {26.78}  &   {133.13\phantom{0}}\\
      
				&\textcolor{orange}{$\bf\diamond$} \textbf{MDDA-former}
				& \underline{33.21}     & \phantom{0}67.38   &25.92  & 32.75 \\

				\hline
				
				
				
				
				

				\multirow{5}{*}{{DND}}  
				


                    &\textcolor{green}{$\bf\diamond$} ADFNet \cite{ADFNet}
				& 39.87              & \phantom{0}55.64   &\phantom{0}7.65  & 48.87 \\
                    
				&\textcolor{orange}{$\bf\diamond$} MAXIM-3S \cite{MAXIM}
				& 39.84              & 320.32  &20.60  & 59.35\\	
				
				&\textcolor{orange}{$\bf\diamond$} Restormer \cite{Restormer}
				& {40.03}  & 140.99  &26.13  & 81.76 \\


                    &\textcolor{orange}{$\bf\diamond$}  {MambaIR \cite{MambaIR}}
				&  {\underline{40.04}}  &  {229.80}   & {26.78}  &  {133.13\phantom{0}} \\
				
				&\textcolor{orange}{$\bf\diamond$} \textbf{MDDA-former}
				& \textbf{40.05}     & \phantom{0}67.38   &25.92  & 32.75\\
				
				\hline
				
				\multirow{5}{*}{{SOTS}}	 
				
				&\textcolor{green}{$\bf\diamond$} DeHamer\cite{DeHamer}
				& {36.63}            & \phantom{0}48.93    &132.45\phantom{0} &16.45 \\
    				
				&\textcolor{green}{$\bf\diamond$} DehazeFormer-L \cite{RS-Haze}
				& {40.05}  & 279.70    &25.44	&54.32	\\

                &\textcolor{green}{$\bf\diamond$} MITNet \cite{MITNet}  
				& {40.23}  & \phantom{0}16.42   &\phantom{0}2.73	&39.38	\\
				
				

                &\textcolor{orange}{$\bf\diamond$}  {MambaIR \cite{MambaIR}}
				&  {\underline{40.58}}  &  {229.80}   & {26.78}  &  {133.13\phantom{0}} \\
				
				&\textcolor{orange}{$\bf\diamond$} \textbf{MDDA-former}
				& \textbf{40.79}     & \phantom{0}67.38    &25.92  &32.75 \\
				
				\hline
				\multirow{5}{*}{{LOL-v1}} 
				&\textcolor{green}{$\bf\diamond$} URetinex-Net \cite{URetinex-Net}
				& {21.33}            & \phantom{0}56.93   &\phantom{0}0.40   & 10.97\\
				
    

                &\textcolor{green}{$\bf\diamond$} UHDFour \cite{UHDFour}
				& {23.09}  & -   &17.54  &\phantom{0}9.68 \\

                    &\textcolor{orange}{$\bf\diamond$} MIRNet \cite{MIRNet}
				& \underline{24.14}  & 789.39  &31.79  & 70.68 \\

                &\textcolor{orange}{$\bf\diamond$} {MambaIR \cite{MambaIR}}
				&  {24.05}  &  {229.80}   & {26.78}  &  {133.13\phantom{0}}  \\
                
				&\textcolor{orange}{$\bf\diamond$} \textbf{MDDA-former}
				& \textbf{24.27}     & \phantom{0}67.38   &25.92  & 32.75\\
				
				\bottomrule[1pt]
		\end{tabular}}
		\vspace{-2.5em}
	\end{center}
\end{table}

\begin{figure*}[!htp]  
	\centerline{\includegraphics[page=1,trim = 0mm 0mm 0mm 0mm, clip, width=1\linewidth]{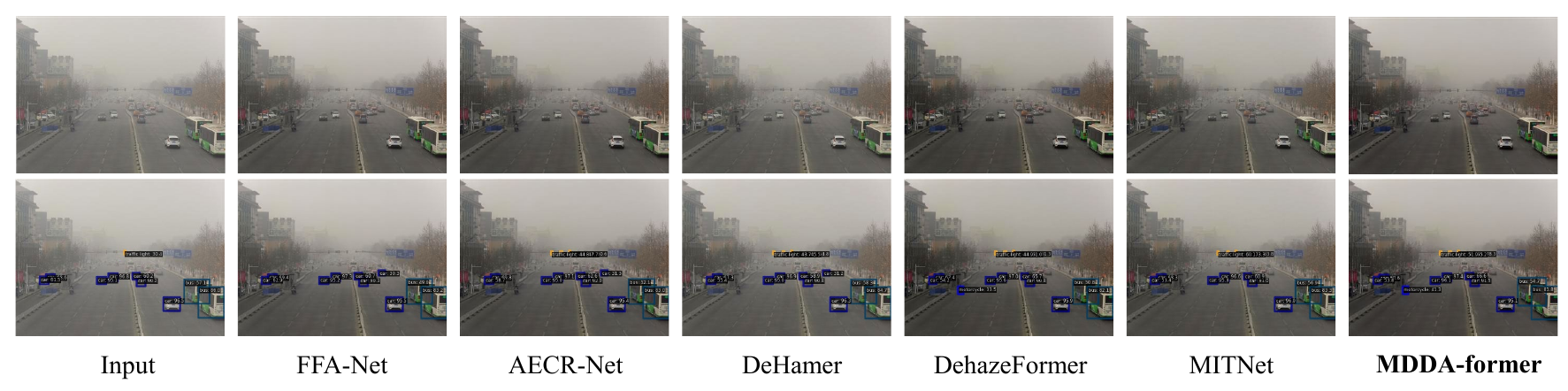}}
\vspace{-1em} 
	\caption{{Visual results of joint dehazing and detection on the RESIDE-RTTS dataset. All the dehazing methods are trained on RESIDE-OTS dataset. The first row denotes the dehazing results and the second row is the results of detection by YOLOv3 backbone.} }
	\label{fig:detection}
	\vspace{-0.5em}
\end{figure*}

\begin{figure*}[!htp]  
	\centerline{\includegraphics[page=1,trim = 0mm 0mm 0mm 0mm, clip, width=1\linewidth]{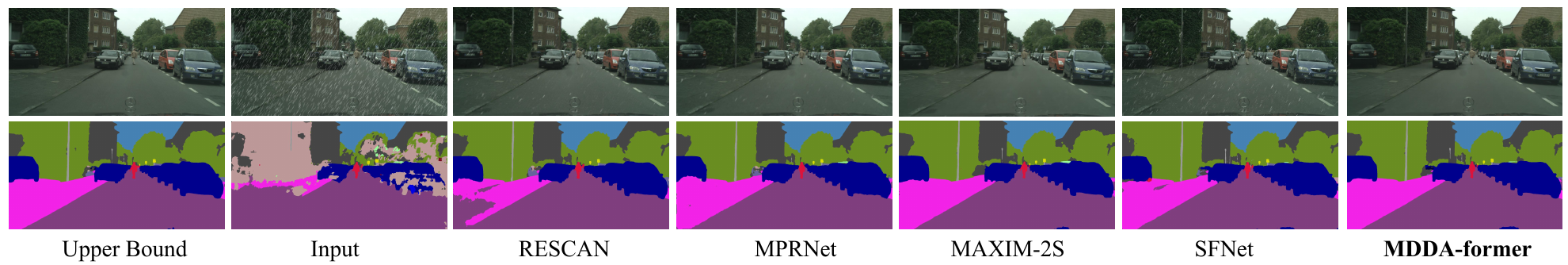}}
\vspace{-1em} 
	\caption{Visual results of joint deraining and segmentation on the Cityscapes validation dataset. All the deraining methods are trained on Rain13K dataset. The first row denotes the deraining results and the second row is the results of semantic segmentation by PSPNet backbone. }
	\label{fig:segmetation}
	\vspace{-1em}
\end{figure*}

\begin{table*}[!h]
\setlength{\abovecaptionskip}{2pt}
	\caption{{Comparison of haze removal and object detection results with the metric as mAP@50 on the RTTS dataset. mAP@0.5 means the mAP calculated at IOU threshold 0.5.}}
	\label{table:detection}
	\begin{center}
		\vspace{-1.0em}
		\renewcommand\arraystretch{1.1}
		
		\scalebox{1}{\begin{tabular}{l|ccccccc}
				\toprule[1pt]
    \toprule[0.5pt]
				
				\backslashbox{\textbf{Backbone}}{\textbf{Method}}
				&{Input}
				&{FFA-Net \cite{FFA-Net}}
				&{AECR-Net \cite{AECR-Net}}
				&{DeHamer \cite{DeHamer}}
                &{DehazeFormer \cite{RS-Haze}}
                &{MITNet \cite{MITNet}}
				&\textbf{MDDA-former}
				\\
				
				\hline

				Faster R-CNN \cite{Faster-RCNN}    & 63.1  & 63.5 & 64.7  & 65.2  & 65.8   & 66.4  &\textbf{66.9} \\
				
				Mask R-CNN \cite{Mask-RCNN} & 64.7  & 64.9 &65.3  & 66.9  & 67.1   &67.2   &\textbf{67.8} \\
				
				YOLOv3 \cite{YOLOv3} & 65.4 &65.9  &65.7   & 66.3  &67.2  & 66.8 & \textbf{67.6}\\
				
				\bottomrule[1pt]
		\end{tabular}}
		\vspace{-1.7em}
	\end{center}
	
\end{table*}

\begin{table*}[!h]
\setlength{\abovecaptionskip}{2pt}
	\caption{ Comparison of rain removal and semantic segmentation results with the metrics such as PSNR, SSIM, and mIoU on the Cityscapes validation dataset. The Upper Bound indicates using clean images to compute segmentation performance and rain removal performance. }
	\label{table:segmentation}
	\begin{center}
		\vspace{-1.0em}
		\renewcommand\arraystretch{1.1}
		
		\scalebox{1}{\begin{tabular}{l|ccccccc}
				\toprule[1pt]
    \toprule[0.5pt]
				
				\backslashbox{\textbf{Backbone}}{\textbf{Method}}
				&{Input}
				&{RESCAN \cite{RESCAN}}
				&{MPRNet \cite{MPRNet}}
				&{MAXIM-2S \cite{MAXIM}}
                &{SFNet \cite{SFNet}}
				&\textbf{MDDA-former}
                & Upper Bound
				\\
				
				\hline
				
				FCN \cite{FCN}       & 31.8  & 61.3 &67.4  & 68.8  & 66.1   & \textbf{70.2}  &73.4 \\
				
				PSPNet \cite{PSPNet} & 33.5  & 63.7 &70.1  & 72.8  & 69.3   &\textbf{73.7}   &78.5 \\
				
				DeepLabv3+ \cite{DeepLabv3+} & 34.3 &65.9  &71.6   & 74.2  &70.9  & \textbf{75.1} &79.8\\
				
\cdashline{1-8}
	
				PSNR$\uparrow$/SSIM$\uparrow$ & 23.14/0.724  & 31.97/0.916 &34.17/0.931 & 34.62/0.938  & 34.07/0.923  &\textbf{35.16}/\textbf{0.949}  &$\infty$/1 \\

				\bottomrule[1pt]
		\end{tabular}}
		\vspace{-1.7em}
	\end{center}
	
\end{table*}

\noindent\textbf{Applying to High-level Vision Tasks.} To verify the potential of our proposed approach in helping improve the performance of high-level vision tasks for intelligent driving, we conduct a series of experiments on downstream tasks, including object detection and semantic segmentation.

For the object detection task,  {we select the RTTS dataset in RESIDE \cite{RESIDE} for detailed illustration, which consists of 4322 real-world outdoor city-scene images with annotations. Three popular detection algorithms are adopted, including Faster R-CNN \cite{Faster-RCNN}, Mask R-CNN \cite{Mask-RCNN}, and YOLOv3 \cite{YOLOv3}. The comparison methods for dehazing were pretrained on the RESIDE-OTS  dataset and validated on the RESIDE-RTTS to restore high quality results.  Then, we adopt the publicly pretrained models\footnote{MMDetection: \url{https://github.com/open-mmlab/mmdetection}} of Faster R-CNN, Mask R-CNN, and YOLOv3 to perform the object detection task on these restored images. The mean Average Precision (mAP) was adopted as the performance metric. 
Tab. \ref{table:detection} shows the object detection results measured with the mAP metric on the restored image after haze removal of the RTTS dataset. It is observed that our proposed MDDA-former achieves the highest results across three detection backbones. A visual example of haze removal and detection among comparison methods is demonstrated in Fig. \ref{fig:detection}. With careful observation, compared to the input, the restored images have all been improved with higher contrast, and more vehicles have been detected. Particularly, the motorcycle has been detected by only DehazeFormer and the proposed MDDA-former, which indicates that our MDDA-former can restore high-quality images that are favorable for object detection tasks against the degraded input and most comparison approaches.}  

 For the semantic segmentation task, we followed RCDNet \cite{RCDNet} and used the Cityscapes validation set \cite{cityscapes} as the benchmark which includes 500 images with annotations. Similar to \cite{RCDNet}, we introduce rain to the Cityscapes validation set with Photoshop. FCN \cite{FCN}, PSPNet \cite{PSPNet}, and DeepLabv3+ \cite{DeepLabv3+} were utilized as the segmentation backbones and corresponding pre-trained models\footnote{MMSegmentation: \url{https://github.com/open-mmlab/mmsegmentation}} are applied to perform the segmentation task on these restored images. The Mean Intersection over Union (mIoU) was adopted as the performance metric. Fig. \ref{fig:segmetation} presents the visual segmentation results of PSPNet on input, upper bound (rain-free), and restored images after rain removal through different methods. It can be observed from the column of input that the presence of rain streaks results in poor segmentation outcomes for the rain significantly degrades the image details. Among comparison methods, our proposed MDDA-former could remove rain more thoroughly and preserve more details, thus leading to higher accuracy of semantic segmentation.  As reported in Tab. \ref{table:segmentation}, it is readily apparent that the PSNR, SSIM, and segmentation accuracy of the derained images restored by our MDDA-former are the highest except for `upper bound' and consistently improve around 40\% against those of rainy images under the segmentation backbones. These observations demonstrate the feasibility of the MDDA-former for applications in high-level vision tasks.

{\noindent\textbf{Applying to Multi-task Scene.} To test whether the proposed MDDA-former could handle multiple degradations at a time, experiments were conducted to evaluate the performance of the proposed MDDA-former and comparison methods across three distinct types of degradations. We employed the same experimental setup as PromptIR\cite{PromptIR}, with the detailed dataset configuration delineated in Tab. \ref{table:Dataset}. Tab. \ref{table:all_in_one} shows the results of three restoration tasks for ``noise-haze-rain'' among the proposed MDDA-former and comparison methods including general image restoration and all-in-one methods. It could be observed that the proposed MDDA-former could achieve better results than the listed general image restoration methods, and even better than the all-in-one method AirNet. {MDDA-former could achieve superior performance with MambaIR \cite{MambaIR} on denoising performance while lagging by over 0.40/0.16 dB on dehazing and deraining.} 
MDDA-former could achieve comparable performance with PromptIR on denoising performance while lagging by over 1 dB on dehazing and deraining.  In all, the proposed MDDA-former could have fairly good results in addressing multiple degradation patterns but is much inferior to the proposed model only trained for one task. Nonetheless, more efforts considering the difference in degradation types and severity should be made for all-in-one restoration methods.}

\begin{table*}[!htbp] 
\setlength{\abovecaptionskip}{1pt}
	\caption{ Quantitative results on \textbf{All-in-One (``noise-haze-rain'')} setting with state-of-the-art general image restoration methods (marked in \textcolor{orange}{$\bf\diamond$}) and all-in-one methods (marked in \textcolor{xx}{$\bf\diamond$}).}  
	\label{table:all_in_one}
		\vspace{-1em}
	\begin{center}
		\renewcommand\arraystretch{1.1}
		\begin{threeparttable}
		\resizebox{\linewidth}{!}{\begin{tabular}{l|c||ccc|c|c|c||c}
					 
                \toprule[1pt]
                \toprule[0.5pt]
        
                \multirow{2}{*}{\textbf{Method}} 
                & \multirow{2}{*}{\textbf{Venue}}
                & \multicolumn{3}{c|}{\textbf{Denoising (CBSD68 \cite{CBSD68})}}
                & \multicolumn{1}{c|}{\textbf{Dehazing}}
                & \multicolumn{1}{c|}{\textbf{Deraining}}
                & \multirow{2}{*}{\textbf{Average}}
                & \multirow{2}{*}{\makecell{\textbf{FLOPs (G)$\downarrow$/}\\\textbf{\#Params (M)$\downarrow$}}}
                \\ \cline{3-5} \cline{6-6} \cline{7-7}  
                        & 
					& $\sigma = 15$ & $\sigma = 25$ & $\sigma = 50$  
					&SOTS-Outdoor \cite{RESIDE}  & Rain100L \cite{Rain100}
					&
					  \\ 
									
					\hline
					\textcolor{orange}{$\bf\diamond$} MPRNet \cite{MPRNet}
                        & 21'CVPR
                        & 33.27/0.920  & 30.76/0.871  & 27.29/0.761
					& 28.00/0.958
					& 33.86/0.958
					& 30.63/0.894 
                        & 148.55/\phantom{0}3.64
					\\
					
					\textcolor{orange}{$\bf\diamond$} Restormer \cite{Restormer}
                        & 22'CVPR
					& 33.72/0.930  & 30.67/0.865  & 27.63/0.792
                        & 27.78/0.958
					& 33.78/0.958 
					& 30.75/0.901
                        & 140.99/26.13
					\\
					
					\textcolor{orange}{$\bf\diamond$} NAFNet \cite{NAF}
                        & 22'ECCV
					& 33.03/0.918  & 30.47/0.865  & 27.12/0.754
                        & 24.11/0.928
					& 33.64/0.956
					& 29.67/0.844
                        & \phantom{0}63.33/67.89
					\\

					\textcolor{orange}{$\bf\diamond$} {MambaIR \cite{MambaIR}}
                        & {24'ECCV}
					& {33.88/0.931}  & {30.95/0.874}  & {27.74/0.793}
                        & {29.57/0.970}
					& {35.42/0.969}
					& {31.51/0.907}
                        & {229.80/26.78}
					\\

                        \textcolor{orange}{$\bf\diamond$} \cellcolor{mycolor}\textbf{MDDA-former}
                        & \cellcolor{mycolor}-
					& \cellcolor{mycolor}33.94/0.933  & \cellcolor{mycolor}31.28/0.888  &  
                        \cellcolor{mycolor}28.05/0.798
                        & \cellcolor{mycolor}29.17/0.968
					& \cellcolor{mycolor}35.26/0.969
					& \cellcolor{mycolor}\underline{31.54}/\underline{0.911}
                        & \cellcolor{mycolor}\phantom{0}67.38/25.92
					\\
     
                        \cdashline{1-9}
                        \textcolor{xx}{$\bf\diamond$} AirNet \cite{AirNet}
                        & 22'CVPR
                        & 33.92/0.933  & 31.26/0.888  & 28.00/0.797  
                        & 27.94/0.962   & 34.90/0.967  & 31.20/0.910
                        & 301.27/\phantom{0}5.77
					\\

                        \textcolor{xx}{$\bf\diamond$} PromptIR \cite{PromptIR}
                        & 23'NeurIPS
                        & 33.98/0.933  & 31.31/0.888  & 28.06/0.799    
                        & 30.58/0.974  & 36.37/0.972  & \textbf{32.06}/\textbf{0.913}
                        & 158.14/32.96
					\\

					\bottomrule[1pt]
			\end{tabular}}
		\end{threeparttable}
	\end{center}
	\vspace{-2.3em}
\end{table*}

\subsection{Ablation Study} 
In this section, a relatively smaller version MDDA-former* was introduced to efficiently conduct the ablation study, in which the feature dimensions at each stage were set as [48, 96, 192, 384, 192, 96, 48] and the number of MDABs in the encoder-decoder was set as [2, 6, 8, 4, 3, 2]. We explore the impacts of each component and different architectures of MDDA-former by analyzing these effects in MDDA-former*. The evaluations are conducted on image deraining (Rain13K \cite{MSPFN}) and dehazing (RESIDE-ITS \cite{RESIDE}) tasks. For simplicity, the models were trained for 200K iterations and performed on Rain100L \cite{Rain100} and SOTS-Indoor \cite{RESIDE} datasets. The other settings were the same as MDDA-former except for special instructions.

\emph{1) Effects of attention mechanism in the MDAB}: To validate the effectiveness of our designed MDConv, the model with MDConv was compared with the models with other attention mechanisms within the MDAB, such as CE~\cite{SE}, CBAM~\cite{CBAM}, ECA~\cite{ECA}, DyConv~\cite{DyConv}, and ODConv~\cite{ODConv}. In ODConv, $n$ represents the number of convolutional kernels.
As shown in Tab. \ref{table:ablation_mechanism}, although models e and f achieve the higher PSNR values on Rain100L dataset~\cite{Rain100}, ODConv also introduces a large number of additional model parameters than the other attentions within the MDAB. Furthermore, the MDAB equipped with MDConv outperforms the MDAB with all other attentions except for ODConv in terms of PSNR on SOTS-Indoor dataset~\cite{RESIDE}. This demonstrates that MDConv strikes a balance between performance and complexity.

		
				
				
				
				
				
				
				
				
				
	


{\emph{2) Effects of the attention components in the MDConv}: To further explore the effect of individual alpha (attention) in the MDConv, the ablation studies have also been conducted on the derain task. As illustrated in Tab. \ref{table:ablation_alpha}, from cases (a)-(c), where only one type of alpha is used, it can be inferred that $\bm{\alpha}_{c}$ (channel-wise attention) plays the most significant role in the MDConv, $\bm{\alpha}_{s}$ (spatial-wise attention) the second, and $\bm{\alpha}_{f}$ (filter-wise attention) the third. From cases (d)-(f), where two attentions out of all three alphas are combined and employed, it can be found that the combination of $\bm{\alpha}_{c}$ and $\bm{\alpha}_{s}$ is superior to the other two cases of combinations of two alphas. By contrast, the proposed method, i.e., using all alphas, achieves the best among all the other cases.}

\emph{3) Effects of number of ETB}: We study the influence of the number of ETB in Tab. \ref{table:number_ETB}. As can be seen, using more ETBs leads to the consistently increasing performance from 37.31 to 37.53 dB PSNR for the deraining task and 37.65 to 37.87 dB PSNR for the dehazing task, while introducing 6.3 M parameters and 6.45 G FLOPs. To achieve the performance and efficiency trade-off, we choose to set the number of ETBs to 10, as increasing it to 12 only slightly improves performance but adds more complexity. 

{\emph{4) Effects of learning parameters in ETB}: To explore the effects of the learnable parameters $\mathbf{k}_{1}$ and $\mathbf{k}_{2}$, an ablation study was conducted on the derain task. As illustrated in Tab. \ref{table:learnable}, the proposed method with both $\mathbf{k}_{1}$ and $\mathbf{k}_{2}$ achieves the best performance while the proposed method without $\mathbf{k}_{1}$ and $\mathbf{k}_{2}$ slightly deteriorates the performance. The role of learning parameters likely contributes to stabilizing the training process.}  

{\emph{5) Effects of the chunk operation}: As shown in Tab. \ref{table:ablation_thechunk}, using the chunk operation in both ETB and MDAB (i.e., ours) slightly reduces computational complexity but hurts PSNR performance a little as compared to the scheme not using it in both ETB and MDAB (i.e., model a). In addition, using the chunk operation in either MDAB or ETB (i.e., models b or c) also reduces computational complexity, but the performance reduction is more significant than when using it in both MDAB and ETB (i.e., model a). Overall, using chunk operation in MDDA-former is a trade-off between performance and complexity.}

\begin{table}[!tbp]
\setlength{\abovecaptionskip}{2pt}
	\caption{Exploring the effects of different attentions within the MDAB.}
	\label{table:ablation_mechanism}
	\begin{center}
		\vspace{-1.5em}
		\renewcommand\arraystretch{1.1}	
		\resizebox{\linewidth}{!}{\begin{tabular}{l|cc|c}
				\toprule[1pt]
				\textbf{Attention}
				&\textbf{FLOPs (G)}
				&\textbf{\#Params (M)}
				&\textbf{PSNR$\uparrow$ (rain/haze)}
				\\
				
				\hline
				SE \cite{SE} &39.50 & 12.43  &36.49/36.63\\   
				CBAM \cite{CBAM} & 39.59  &12.43 &36.33/36.49\\ 
				ECA \cite{ECA} & 39.50  &12.36 & 36.29/36.47 \\  
				DyConv \cite{DyConv} & 39.50  &16.27 &36.53/36.81  \\  
				ODConv ($n$=2) \cite{ODConv} & 39.50  &26.39 &\underline{37.62}/37.69 \\  
				ODConv ($n$=4) \cite{ODConv} & 39.50  &36.26 &\textbf{37.71}/\underline{37.82} \\  
				\cellcolor{mycolor}\textbf{MDConv (Ours)} 
                & \cellcolor{mycolor}39.50 
                & \cellcolor{mycolor}16.49  
                & \cellcolor{mycolor}37.49/\textbf{37.86} \\   
				
				\bottomrule[1pt]
		\end{tabular}}
		\vspace{-1.7em}
	\end{center}
\end{table}

\begin{table}[!tbp]
	\caption{
 Exploring the effects of different attention components in the MDConv.}\label{table:ablation_alpha}
	\begin{center}
		\vspace{-1.5em}
		\renewcommand\arraystretch{1.1}
		
		\scalebox{1}{\begin{tabular}{c|ccc|c}
				\toprule[1pt]
				
				\textbf{Model}
				&${\bm{\alpha}_{s}}$
				&${\bm{\alpha}_{c}}$
				&${\bm{\alpha}_{f}}$
				&\textbf{PSNR$\uparrow$ (rain)}
				\\
				\hline
				
				(a) 
				& \textcolor{green}{\ding{51}}   
                &\textcolor{red}{\ding{55}}   
                & \textcolor{red}{\ding{55}}    
                    &36.71
				\\
				
				(b)
				&\textcolor{red}{\ding{55}}  
                &\textcolor{green}{\ding{51}} 
                &\textcolor{red}{\ding{55}}     
                    &36.81
				\\
				
				(c)
				&\textcolor{red}{\ding{55}}   
                &\textcolor{red}{\ding{55}} 
                &\textcolor{green}{\ding{51}}                         
				&36.65       
                    \\
                    
                \cdashline{1-5}
                (d)
				&\textcolor{green}{\ding{51}}   
                &\textcolor{green}{\ding{51}} 
                &\textcolor{red}{\ding{55}}                      
				&37.14       
                    \\

                    (e)
				&\textcolor{green}{\ding{51}}   
                &\textcolor{red}{\ding{55}} 
                &\textcolor{green}{\ding{51}}                      
				&36.97       
                    \\

                    (f)
				&\textcolor{red}{\ding{55}}   
                &\textcolor{green}{\ding{51}} 
                &\textcolor{green}{\ding{51}}                      
				&37.05       
                    \\
                    
                \cdashline{1-5}
				\cellcolor{mycolor}(\textbf{Ours})
				&\cellcolor{mycolor}\textcolor{green}{\ding{51}}  &\cellcolor{mycolor}\textcolor{green}{\ding{51}} 
                & \cellcolor{mycolor}\textcolor{green}{\ding{51}}
				&\cellcolor{mycolor}\bf37.49       
                \\
				
				\bottomrule[1pt]
		\end{tabular}}
		\vspace{-1.7em}
	\end{center}
	
\end{table}

\begin{table}[!tbp]
	\caption{Exploring the effects of the number of ETB.}
	\label{table:number_ETB}
	\begin{center}
		\vspace{-1.2em}
		\renewcommand\arraystretch{1.1}
		
		\scalebox{1}{\begin{tabular}{c|cc|c}
				\toprule[1pt]
				
				\textbf{Number}
                &\textbf{FLOPs (G)}
				&\textbf{\#Params (M)}
                &\textbf{PSNR$\uparrow$ (rain/haze)}
				\\
				\hline

				6  & 35.20 & 12.29 & 37.31/37.65 \\
				
				8  &37.25  & 14.39 & 37.38/37.73 \\
				
				\cellcolor{mycolor}\textbf{10}  &\cellcolor{mycolor}39.50 & \cellcolor{mycolor}16.49 & \cellcolor{mycolor}37.49/37.86  \\
				
				12  &41.65 & 18.59   & 37.53/37.87  \\ 
					
				\bottomrule[1pt]
		\end{tabular}}
		\vspace{-2.3em}
	\end{center}
	
\end{table}

\begin{table}[!tbp]
	\caption{Exploring the effects of the learnable parameters in ETB.}\label{table:learnable}
	\begin{center}
		\vspace{-1.5em}
		\renewcommand\arraystretch{1.1}
		
		\scalebox{1}{\begin{tabular}{c|cc|c}
				\toprule[1pt]
				
				\textbf{Model}
				&${\mathbf{k}_{1}}$
				&${\mathbf{k}_{2}}$
				&\textbf{PSNR$\uparrow$ (rain)}
				\\
				\hline
 
				(a) 
				& \textcolor{green}{\ding{51}}   
                &\textcolor{red}{\ding{55}}      
                    &37.41
				\\
				
				(b)
				&\textcolor{red}{\ding{55}}  
                &\textcolor{green}{\ding{51}}      
                    &37.46
				\\

				\cellcolor{mycolor}(\textbf{Ours})
				&\cellcolor{mycolor}\textcolor{green}{\ding{51}}  &\cellcolor{mycolor}\textcolor{green}{\ding{51}} 
				&\cellcolor{mycolor}\bf37.49       
                \\
				
				\bottomrule[1pt]
		\end{tabular}}
		\vspace{-1.7em}
	\end{center}
	
\end{table}

\begin{table}[!tbp]
	\caption{Exploring the effects of the chunk operation.}\label{table:ablation_thechunk}
	\begin{center}
		\vspace{-1.5em}
		\renewcommand\arraystretch{1.1}
		
		\resizebox{\linewidth}{!}{\begin{tabular}{c|cc|cc|c}
				\toprule[1pt]
				
				\textbf{Model}
				&\makecell{chunk in\\ MDAB}
				&\makecell{chunk in\\ ETB}
				&\textbf{FLOPs (G)}
				&\textbf{\#Params (M)}
				&\textbf{PSNR$\uparrow$ (rain)}
				\\
				\hline
				
				(a) 
				&\textcolor{red}{\ding{55}}     &\textcolor{red}{\ding{55}}
				&43.15      &17.29
				&37.56       
				\\
				(b)
				&\textcolor{green}{\ding{51}}  &\textcolor{red}{\ding{55}}
				&42.32      &16.76
				&37.42      
				\\
				
				(c)
				&\textcolor{red}{\ding{55}}        &\textcolor{green}{\ding{51}}
				&40.86      &16.63
				&37.33       
                \\

				\cellcolor{mycolor}(\textbf{Ours})
				&\cellcolor{mycolor}\textcolor{green}{\ding{51}}  &\cellcolor{mycolor}\textcolor{green}{\ding{51}}
				&\cellcolor{mycolor}39.50     &\cellcolor{mycolor}16.49
				&\cellcolor{mycolor}37.49       
                \\

				\bottomrule[1pt]
		\end{tabular}}
		\vspace{-2.3em}
	\end{center}
	
\end{table}

\begin{table*}[!htbp]
\setlength{\abovecaptionskip}{2pt}
	\caption{Exploring the effects of ETB, MDAB, and other modules at encoder-decoder or latent layers within the MDDA-former*.}
	\label{table:ablation_architecture}
	\begin{center}
		\renewcommand\arraystretch{1.1}
  \vspace{-1.5em}
		\resizebox{\linewidth}{!}{\begin{tabular}{c|c|cc|cc||c}
				\toprule[1pt]
                \toprule[0.5pt]

                \textbf{Model}
				&\textbf{Type}
				&\textbf{Encoder-Decoder}
				&\textbf{Latent}
				&\textbf{FLOPs (G)}
				&\textbf{\#Params (M)}
				&\textbf{PSNR$\uparrow$ (rain/haze)}
				\\
    
				\hline
				(a)
                & T-T-T
				& ETB &ETB
				& 59.11 &14.13
				&37.18/37.51 \\
				
				(b)
                & T-C-T
				& ETB &MDAB
				& 52.99 &20.58
				& 36.47/36.92 \\

				
				(c)
                & C-C-C
				& MDAB &MDAB
				& 33.38 &22.96
				&{37.39/37.64} \\
							
				(d)
                & C-H-C
				& MDAB &MDAB+ETB (Concat)
				& 35.69  &16.40
				& 37.31/37.63 \\ 

				(e)
                & C-H-C
				& MDAB &MDAB+ETB (Adaptive fusion \cite{Fusion}) 
                &34.93   &16.11
				&37.29/37.59 \\

				(f)
                & H-H-H
				& {MDAB+ETB (Concat)} &{MDAB+ETB (Concat)}
				& 43.24 &13.98
				&37.24/37.55 \\
				
				(g)
                & H-H-H
				&{MDAB+ETB (Adaptive fusion \cite{Fusion})} &{MDAB+ETB (Adaptive fusion \cite{Fusion})}
				&40.08  &13.61
				&37.18/37.46 \\

				(h)
                & T-T-T
				& MDTA+GDFN \cite{Restormer} & ETB
				& 74.01 &20.03
				&37.27/37.69 \\
				
				(i)
                & T-T-T
				& W-MSA+LeFF \cite{Uformer} & ETB
				& 78.23 &36.46
				& 37.38/37.73 \\
						
				(j)
                & C-T-C
				& MDAB  & MDTA+GDFN \cite{Restormer}
				& 44.15 &20.98
				&{37.36/37.75} \\

				(k) 
                & C-T-C
				& MDAB &W-MSA+LeFF \cite{Uformer}
				& 51.84  &27.16
				&{37.25}/{37.64} \\

				\cellcolor{mycolor}(\textbf{Ours})
                & \cellcolor{mycolor}C-T-C
				& \cellcolor{mycolor}MDAB &\cellcolor{mycolor}ETB
				& \cellcolor{mycolor}39.50  &\cellcolor{mycolor}16.49
				&\cellcolor{mycolor}\textbf{37.49}/\textbf{37.86} \\

				
				\bottomrule[1pt]
		\end{tabular}}
		\vspace{-1.5em}
	\end{center}
\end{table*}

\begin{figure*}[!tp] 
	\centerline{\includegraphics[page=1,trim = 0mm 0mm 0mm 0mm, clip, width=1\linewidth]{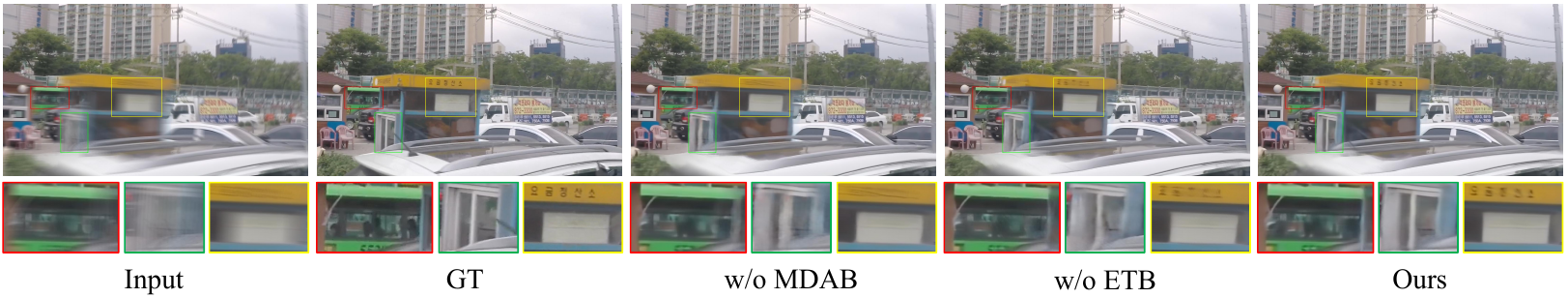}}
   \captionsetup{skip=0pt}
	\caption{Qualitative ablation studies of the MDAB and ETB.}
	\label{fig:ablation}
	\vspace{-1.5em}  
\end{figure*}

\emph{6) Discussion on the role of model architecture}: To explore the impacts of the model architecture and proposed modules in the MDDA-former, different combinations and allocations of MDAB and ETB are explored within the MDDA-former*. As shown in Tab. \ref{table:ablation_architecture}, first, we fixed one module and changed the other module, i.e., models a and c, and exchanged the two modules, i.e., model b, it can be observed that these three cases are all inferior to our model. As demonstrated in  Fig. \ref{fig:ablation}, it could be observed that the scheme `w/o MDAB' (model a) would retain the global structures but lose fine details of bus, windows, and characters while the scheme `w/o ETB' (model c) could add more details. By comparison, our proposed model could perform better both globally and locally.  The objective and the visual results may suggest that the architecture in the MDDA-former* is effective by fully exploiting the advantages and avoiding the weaknesses of the transformer and CNN. 

However, whether the combinations of transformer- and CNN-based modules in the architecture are effective needs to be explored. Then, connecting MDAB and ETB in a concatenation manner (i.e., models d and f) and the simple adaptive fusion of MDAB and ETB (i.e., models e and g) in the latent layer, and both encoder-decoder and latent layer were tested. It can be found that these models are still all worse than our model, which may be because the improper fusion of MDAB and ETB could lead to performance degradation. Finally, whether other existing hybrid architectures of transformer and CNN modules are more effective, MDTA+GDFN \cite{Restormer} and W-MSA+LeFF \cite{Uformer} were deployed at encoder-decoder (i.e., models h and i) or latent layers (i.e., models j and k) respectively within the MDDA-former*.  It can be seen that models h to k do not yield results as effective as our proposed scheme. Though the above validations may not be thorough and rigorous, the findings could still underscore the potency of the proposed ETB and MDAB, and suggest that their judicious deployment can enhance the performance of the MDDA-former.

\vspace{-1.3em} 
\subsection{Limitations and Future Work}
Although the MDDA-former, with its carefully designed architecture, has a lower number of parameters or FLOPs than using only MDAB or ETB on the entire U-shaped architecture, it may not be the optimal option to directly stack MDAB and ETB in the order of the U-shaped structure. Developing an adaptive fusion module may be a better option \cite{Fusion}, however, creating an appropriate adaptive fusion module for integrating MDAB and ETB within our model remains a challenge. As indicated in Tab. \ref{table:ablation_architecture}, simply using the adaptive fusion module to combine ETB and MDAB corrupted the model performance as compared with the direct way in the proposed scheme. Also, the inference speed of the proposed MDDA-former is not efficient enough to realize the real-time image processing for intelligent driving with access to weather conditions, thus more lightweight design needs to be explored.
{Moreover, since our method is not specifically designed as an all-in-one restoration solution, the performance of the universal weather removal task could be further improved by incorporating distortion type-aware and distortion level-aware designs leveraging the formidable capabilities of large-scale language-vision models.}


\vspace{-0.5em} 
\section{Conclusions}
In this paper, we have proposed an effective and efficient U-shaped hybrid transformer network (termed  MDDA-former) for general image restoration, which combines the merits of CNNs for local modeling and Transformers for global modeling. Our U-shaped architecture incorporates Efficient Transformer Blocks (ETBs) and CNN-based Multi-dimensional Dynamic Attention Blocks (MDABs) into the latent layer and the encoder-decoder, respectively. This design capitalizes on the global and local modeling capabilities of transformers and CNNs, while circumventing high computational complexity. Specifically, the MDAB is engineered to effectively extract local contextual cues from degraded areas by employing dynamic complementary attentions along the spatial, channel, and filter dimensions of the convolutional kernel. Concurrently, the ETB is designed to capture long-distance dependency from high-level latent features.  Extensive experiments have been conducted on 18 benchmark datasets across five image restoration tasks, demonstrating that the proposed MDDA-former either outperforms or matches the performance of relevant state-of-the-art methods. 
\vspace{-0em} 



%

%
%
%
%
%

\ifCLASSOPTIONcaptionsoff
  \newpage
\fi

\bibliographystyle{IEEEtran}
\bibliography{IEEEabrv,MDDA_former_arxiv}

\end{document}